%% file: main.tex
\documentclass[10pt,twocolumn,letterpaper]{article}
\usepackage{cvpr}
\usepackage{times}
\usepackage{epsfig}
\usepackage{graphicx}
\usepackage{amsmath}
\usepackage{amssymb}
\usepackage{multirow}
\usepackage{subcaption}
\usepackage[linesnumbered,ruled]{algorithm2e}
\usepackage{pbox}
\usepackage[skip=2pt]{caption}
\usepackage{makecell}
\usepackage[export]{adjustbox}    
\usepackage{diagbox}             % added
\usepackage{cuted}
\usepackage{capt-of}

\usepackage[breaklinks=true,bookmarks=false]{hyperref}
\cvprfinalcopy % *** Uncomment this line for the final submission

\def\cvprPaperID{593} % *** Enter the CVPR Paper ID here

\begin{document}
	
	\title{3DN: 3D Deformation Network}
	
	\author{Weiyue Wang$^1$ \qquad Duygu Ceylan$^2$
		\qquad Radomir Mech$^2$  \qquad Ulrich Neumann$^1$\\
		\hspace{-15mm}$^1$University of Southern California \hspace{30mm} $^2$Adobe\\
		\hspace{5mm}Los Angeles, California \hspace{35mm} San Jose, California\\
		{\tt\small \hspace{6mm}\{weiyuewa,uneumann\}@usc.edu}\hspace{28mm}{\tt\small \{ceylan,rmech\}@adobe.com}\qquad
	}

	\maketitle
	
	\vspace{-10pt}
	\begin{abstract}
		\input{abstract}
	\end{abstract}
	
	\input{introductionNew}
	\input{relatedWork}
	\input{method}
	\input{evaluation}
	\input{conclusion}

	{\small
		\bibliographystyle{ieee}
		\bibliography{egbib}
	}
	
\end{document}

%% file: abstract.tex
Applications in virtual and augmented reality create a demand for rapid creation and easy access to large sets of 3D models. An effective way to address this demand is to edit or deform existing 3D models based on a reference, e.g., a 2D image which is very easy to acquire. Given such a source 3D model and a target which can be a 2D image, 3D model, or a point cloud acquired as a depth scan, we introduce \emph{3DN}, an end-to-end network that deforms the source model to resemble the target. Our method infers per-vertex offset displacements while keeping the mesh connectivity of the source model fixed. We present a training strategy which uses a novel differentiable operation, \emph{mesh sampling operator}, to generalize our method across source and target models with varying mesh densities. \emph{Mesh sampling operator} can be seamlessly integrated into the network to handle meshes with different topologies. Qualitative and quantitative results show that our method generates higher quality results compared to the state-of-the art learning-based methods for 3D shape generation. Code is available at \url{github.com/laughtervv/3DN}.

%% file: introductionNew.tex
\section{Introduction}
\label{sec:intro}
Applications in virtual and augmented reality and robotics require rapid creation and access to a large number of 3D models. Even with the increasing availability of large 3D model databases~\cite{chang2015shapenet}, the size and growth of such databases pale when compared to the vast size of 2D image databases. As a result, the idea of editing or deforming existing 3D models based on a reference image or another source of input such as an RGBD scan is pursued by the research community.

Traditional approaches for editing 3D models to match a reference target rely on optimization-based pipelines which either require user interaction~\cite{xu_sig11} or rely on the existence of a database of segmented 3D model components~\cite{Huang:2015:SRV}. The development of 3D deep learning methods~\cite{qi2017pointnet,choy20163d,wu20153d,wang2018sgpn,huang2018recurrent} inspire more efficient alternative ways to handle 3D data. In fact, a multitude of approaches have been presented over the past few years for 3D shape generation using deep learning. Many of these, however, utilize voxel~\cite{Yan2016,Girdhar16b,zhu2017rethinking,marrnet,drcTulsiani17,wu2018,Yang_2018_ECCV,wang2017shapeinpainting} or point based representations~\cite{fan2017point} since the representation of meshes and mesh connectivity in a neural network is still an open problem. The few recent methods which do use mesh representations make assumptions about fixed topology~\cite{groueix2018,wang2018pixel2mesh} which limits the flexibility of their approach.

\begin{figure}
	\begin{center}
		\includegraphics[width=.5\textwidth]{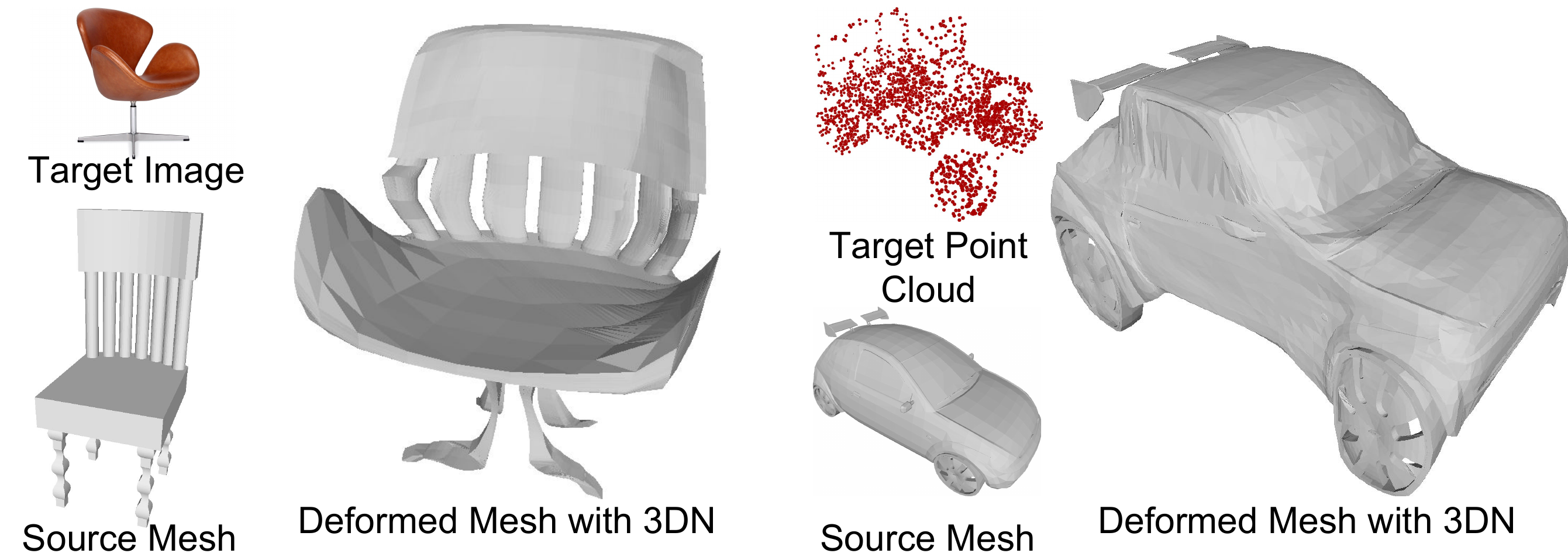}
	\end{center}
	\caption{3DN deforms a given a source mesh to a new mesh based on a reference target. The target can be a 2D image or a 3D point cloud.}
	\label{fig:teaser}
\end{figure}

This paper describes \emph{3DN}, a 3D deformation network that deforms a source 3D mesh based on a target 2D image, 3D mesh, or a 3D point cloud (e.g., acquired with a depth sensor). Unlike previous work which assume a fixed topology mesh for all examples, we utilize the mesh structure of the source model.  This means we can use any existing high-quality mesh model to generate new models. Specifically, given any source mesh and a target, our network estimates vertex displacement vectors (3D offsets) to deform the source model while maintaining its mesh connectivity. In addition, the global geometric constraints exhibited by many man-made objects are explicitly preserved during deformation to enhance the plausibility of the output model.

Our network first extracts global features from both the source and target inputs. These are input to an \emph{offset decoder} to estimate per-vertex offsets. Since acquiring ground truth correspondences between the source and target is very challenging, we use unsupervised loss functions (e.g., Chamfer and Earth Mover's distances) to compute the similarity of the deformed source model and the target. A difficulty in measuring similarity between meshes is the varying mesh densities across different models. Imagine a planar surface represented by just 4 vertices and 2 triangles as opposed to a dense set of planar triangles. Even though these meshes represent the same shape, vertex-based similarity computation may yield large errors. To overcome this problem, we adopt a point cloud intermediate representation. Specifically, we sample a set of points on both the deformed source mesh and the target model and measure the loss between the resulting point sets. This measure introduces a differentiable mesh sampling operator which propagates features, e.g., offsets, from vertices to points in a differentiable manner.

We evaluate our approach for various targets including 3D shape datasets as well as real images and partial points scans. Qualitative and quantitative comparisons demonstrate that our network learns to perform higher quality mesh deformation compared to previous learning based methods. We also show several applications, such as shape interpolation. In conclusion, our contributions are as follows:
\begin{itemize}
    \item We propose an end-to-end network to predict 3D deformation. By keeping the mesh topology of the source fixed and preserving properties such as symmetries, we are able to generate plausible deformed meshes.
    \item We propose a differentiable mesh sampling operator in order to make our network architecture resilient to varying mesh densities in the source and target models.
\end{itemize}

%% file: relatedWork.tex
%------------------------------------------------------------------------
\section{Related Work}
\label{sec:relatedwork}
%-------------------------------------------------------------------------
\subsection{3D Mesh Deformation}
3D mesh editing and deformation has received a lot of attention from the graphics community where a multitude of interactive editing systems based on preserving local Laplacian properties~\cite{Sorkine:2004} or more global features~\cite{gsmc_iwires_sig_09} have been presented. With easy access to growing 2D image repositories and RGBD scans, editing approaches that utilize a reference target have been introduced. Given source and target pairs, such methods use interactive~\cite{xu_sig11} or heavy processing pipelines~\cite{Huang:2015:SRV} to establish correspondences to drive the deformation. The recent success of deep learning has inspired alternative methods for handling 3D data. Yumer and Mitra\cite{yumer2016learning} propose a volumetric CNN that generates a deformation field based on a high level editing intent. This method relies on the existence of model editing results based on semantic controllers. Kurenkov et al. present DeformNet~\cite{kurenkov2017deformnet} which employs a free-form deformation (FFD) module as a differentiable layer in their network. This network, however, outputs a set of points rather than a deformed mesh.Furthermore, the deformation space lacks smoothness and points move randomly. Groueix et al.~\cite{groueix2018b} present an approach to compute correspondences across deformable models such as humans. However, they use an intermediate common template representation which is hard to acquire for man-made objects. Pontes et al.~\cite{pontes2017image2mesh} and Jack et al. ~\cite{jack2018learning} introduce methods to learn FFD. Yang et al. propose Foldingnet~\cite{yang2018foldingnet} which deforms a 2D grid into a 3D point cloud while preserving locality information. Compared to these existing methods, our approach is able to generate higher quality deformed meshes by handling source meshes with different topology and preserving details in the original mesh. %deforms a source mesh while keeping its topology fixed and thus preserves details in the original mesh.

%------------------------------------------------------------------------
\subsection{Single View 3D Reconstruction}
Our work is also related to single-view 3D reconstruction methods which have received a lot of attention from the deep learning community recently. These approaches have used various 3D representations including voxels~\cite{Yan2016,choy20163d,Girdhar16b,zhu2017rethinking,marrnet,drcTulsiani17,wu2018,Yang_2018_ECCV}, point clouds~\cite{fan2017point}, octrees~\cite{ogn2017,hspHane17,wang2018adaptive}, and primitives~\cite{zou20173d,niu_cvpr18}. Sun et al.~\cite{pix3d} present a dataset for 3D modeling from single-images. However, pose ambiguity and artifacts widely occur in this dataset. More recently, Sinha et al.~\cite{Sinha2017} propose a method to generate the surface of an object using a representation based on geometry images. In a similar approach, Groueix et al.~\cite{groueix2018} present a method to generate surfaces of 3D shapes using a set of parametric surface elements. 
The more recent method of Hiroharo et al.~\cite{kato2018renderer} and Kanazawa et al.~\cite{Kanazawa:animal} also uses differentiable renderer and per-vertex displacements as a deformation method to generate meshes from image sets. Wang et al.~\cite{wang2018pixel2mesh} introduce a graph-based network to reconstruct 3D manifold shapes from input images. These recent methods, however, are limited to generating manifolds and require 3D output to be topology invariant for all examples.  
% However, they require key-points correspondences to be annotated across different images. 

%% file: method.tex
%
\section{Method}
\label{sec:method}

\begin{figure*}
	\begin{center}
		\includegraphics[width=\textwidth]{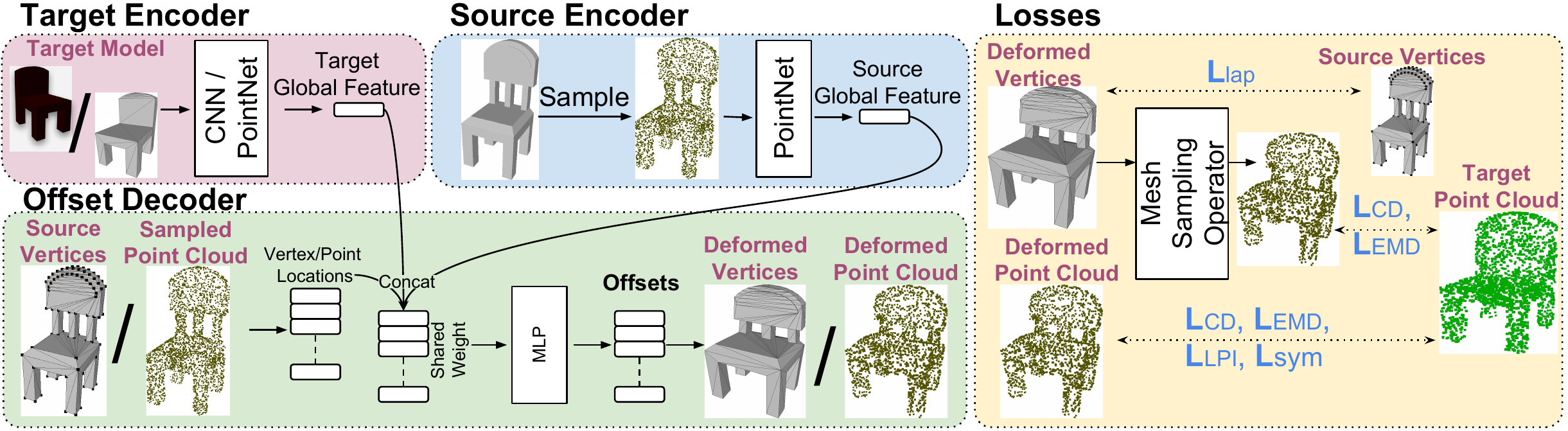}
	\end{center}
	\caption{3DN extracts global features from both the source and target. `MLP' denotes the `$1\times1$' conv as in PointNet~\cite{qi2017pointnet}. These features are then input to an offset decoder which predicts per-vertex offsets to deform the source. We utilize loss functions to preserve geometric details in the source ($L_{Lap}, L_{LPI}, L_{Sym}$) and to ensure deformation output is similar to the target ($L_{CD}, L_{EMD}$).}
	\label{fig:network}
	%\vspace{-10pt}
\end{figure*}

%-------------------------------------------------------------------------
Given a source 3D mesh and a target model (represented as a 2D image or a 3D model), our goal is to deform the source mesh such that it resembles the target model as close as possible. Our deformation model keeps the triangle topology of the source mesh fixed and only updates the vertex positions. We introduce an end-to-end \emph{3D deformation network (3DN)} to predict such per-vertex displacements of the source mesh. 

We represent the source mesh as $S = (V, E)$, where $V \in \mathbb{R}^{N_V \times 3}$ is the $(x,y,z)$ positions of vertices and $E\in \mathbb{Z}^{N_E \times 3}$ is the set of triangles and encodes each triangle with the indices of vertices. $N_V$ and $N_E$ denote the number of vertices and triangles respectively. %For $i^{\text{th}} (0\leq i\le N_E)$ element $\mathbf{e_i}$ in $E$, $\mathbf{e_i} = [p,q,r]$ indicates there is a triangle with vertices $[\mathbf{v_p}, \mathbf{v_q}, \mathbf{v_e}]$ in mesh. 
The target model $T$ is either a $H\times W \times 3$ image or a 3D model. In case $T$ is a 3D model, we represent it as a set of 3D points $T \in \mathbb{R}^{N_T \times 3}$, where $N_T$ denotes the number of points in $T$.% $N_T \times 3$ sampled uniformly on $T$.

As shown in Figure~\ref{fig:network}, 3DN takes $S$ and $T$ as input and outputs per-vertex displacements, i.e., offsets, $O\in \mathbb{R}^{N_V \times 3}$. The final deformed mesh is $S' = (V', E)$, where $V' = V + O$. Moreover, 3DN can be extended to produce per-point displacements when we replace the input source vertices with a sampled point cloud on the source. 3DN is composed of a target and a source encoder which extract global features from the source and target models respectively, and an offset decoder which utilizes such features to estimate the shape deformation. We next describe each of these components in detail.

%-------------------------------------------------------------------------
\subsection{Shape Deformation Network (3DN)}

\paragraph{Source and Target Encoders.}
Given the source model $S$, we first uniformly sample a set of points on $S$ and use the PointNet~\cite{qi2017pointnet} architecture to encode $S$ into a \textit{source global feature vector}. 
Similar to the source encoder, the target encoder extracts a \textit{target global feature vector} from the target model. In case the target model is a 2D image, we use VGG~\cite{simonyan2014very} to extract features. If the target is a 3D model, we sample points on $T$ and use PointNet. We concatenate the source and target global feature vectors into a single \textit{global shape feature vector} and feed into the offset decoder. 
%We note that using a uniformly set of points as opposed to the original model vertices is crucial to make the source encoder prune to different possible mesh topology representing the same geometry. 

\paragraph{Offset Decoder.}
Given the global shape feature vector extracted by the source and target encoders, the offset decoder learns a function $F(\cdot)$ which predicts per-vertex displacements, for $S$. In other words, given a vertex $\mathbf{v}=(x_v, y_v, z_v)$ in $S$, the offset decoder predicts 
$F(\mathbf{v}) = \mathbf{o_v} = (x_{o_v}, y_{o_v}, z_{o_v})$ updating the deformed vertex in $S'$ to be $\mathbf{v'} = \mathbf{v} + \mathbf{o_v}$.

Offset decoder is easily extended to perform point cloud deformations. When we replace the input vertex locations to point locations, say given a point $\mathbf{p}=(x_p, y_p, z_p)$ in the point cloud sampled form $S$, the offset decoder predicts a displacement $F(\mathbf{p}) = \mathbf{o_p}$, and similarly, the deformed point is $\mathbf{p'} = \mathbf{p} + \mathbf{o_p}$.

The offset decoder has an architecture similar to the PointNet segmentation network~\cite{qi2017pointnet}. However, unlike the original PointNet architecture which concatenates the global shape feature vector with per-point features, we concatenate the original point positions to the global shape feature. We find this enables to better capture the vertex and point locations distribution in the source, and results in effective deformation results. We study the importance of this architecture in Section~\ref{sec:ablation}. Finally we note that, our network is flexible to handle source and target models with varying number of vertices.

%. Similarly, if the source mesh is sampled into a point cloud, we can use the same $D(\cdot)$ to compute offset of each point.
%
%The number of input points and output offsets should be the same. And the number of input points can be varying across different samples.

%-------------------------------------------------------------------------
\subsection{Learning Shape Deformations}
Given a deformed mesh $S'$ produced by 3DN and the 3D mesh corresponding to the target model $T = (V_T, E_T)$, where $V_T \in \mathbb{R}^{N_{V_T} \times 3} (N_{V_T} \neq N_V)$ and $E_T \neq E$, the remaining task is to design a loss function that measures the similarity between $S'$ and $T$. Since it is not trivial to establish ground truth correspondences between $S'$ and $T$, our method instead utilizes the Chamfer and Earth Mover's losses introduced by Fan et al.~\cite{fan2017point}. %In order to bridging the gap between deformed mesh $S'$ generated by 3DN and a point set for computing losses inside the network during training, we introduce the \emph{differentiable mesh sampling operator}. 
In order to make these losses robust to different meshing densities across source and target models, we operate on set of points uniformly sampled on $S'$ and $T$ by introducing the \emph{differentiable mesh sampling operator (DMSO)}. DMSO is seamlessly integrated in 3DN and bridges the gap between handling meshes and loss computation with point sets.

\vspace{-5pt}
\paragraph{Differentiable Mesh Sampling Operator.}
\label{sec:meshsampler}

\begin{figure}
%	\hspace{-15pt}
	\begin{center}
		\includegraphics[width=.5\textwidth]{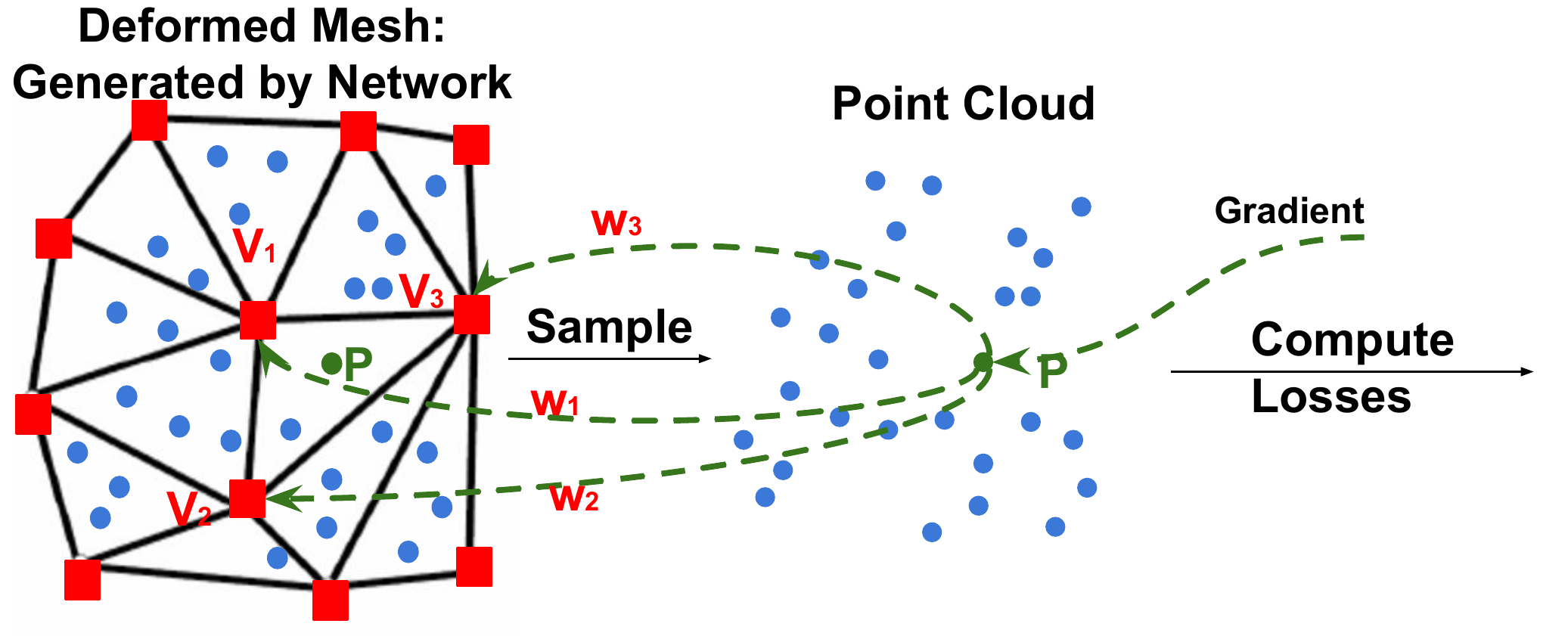}
	\end{center}
	\caption{Differentiable mesh sampling operator (best viewed in color). Given a face $\mathbf{e} = (\mathbf{v_1},\mathbf{v_2},\mathbf{v_3})$, $p$ is sampled on $\mathbf{e}$ in the network forward pass using barycentric coordinates $w_1, w_2, w_3$. Sampled points are used during loss computation. When performing back propagation, gradient of $p$ is passed back to $(\mathbf{v_1},\mathbf{v_2},\mathbf{v_3})$ with the stored weights $w_1, w_2, w_3$. This process is differentiable.}
	\label{fig:meshsampling}
	%\vspace{-10pt}
\end{figure}
As is illustrated in Figure~\ref{fig:meshsampling}, DMSO is used to sample a uniform set of points from a 3D mesh. Suppose a point $\mathbf{p}$ is sampled on the face $\mathbf{e} = (\mathbf{v_1},\mathbf{v_2},\mathbf{v_3})$ enclosed by the vertices $\mathbf{v_1}, \mathbf{v_2}, \mathbf{v_3}$. The position of $\mathbf{p}$ is then 
$$
\mathbf{p} = w_1\mathbf{v_1}+w_2\mathbf{v_2}+w_3\mathbf{v_3},
$$
where $w_1+w_2+w_3=1$ are the barycentric coordinates of $\mathbf{p}$. Given any typical feature for the original vertices, the per-vertex offsets in our case, $\mathbf{o_{v_1}}, \mathbf{o_{v_2}}, \mathbf{o_{v_3}}$, the offset of $\mathbf{p}$ is
$$
\mathbf{o_{p}} = w_{1}\mathbf{o_{v_1}} + w_{2}\mathbf{o_{v_2}} + w_{3}\mathbf{o_{v_3}}.
$$
To perform back-propogation, the gradient for each original per-vertex offsets $\mathbf{o_{v_i}}$ is calculated simply by $g_{\mathbf{o_{v_i}}} = w_i g_{\mathbf{o_{v_p}}}$, where $g$ denotes the gradient.

%\subsubsection{Losses}
%\label{sec:losses}
We train 3DN using a combination of different losses as we discuss next.

\paragraph{Shape Loss.} Given a target model, $T$, inspired by \cite{fan2017point}, we use Chamfer and Earth Mover's distances to measure the similarity between the deformed source and the target. Specifically, given the point cloud $PC$ sampled on the deformed output and $PC_T$ sampled on the target model, Chamfer loss is defined as 
\begin{align}
L_{\texttt{CD}}^{\texttt{Mesh}}(PC, PC_T) = & \sum_{p_1\in PC} \min_{p_2 \in PC_T} \|p_1 - p_2\|_2^2 \nonumber \\ 
+ & \sum_{p_2 \in PC_T} \min_{p_1 \in PC} \| p_1 - p_2 \|_2^2,
\end{align}
and Earth Mover's loss is defined as
\begin{equation}
L_{\texttt{EMD}}^{\texttt{Mesh}}(PC, PC_T) = \min_{\phi: PC \rightarrow PC_T} \sum_{p \in PC} \| p - \phi(p) \|_2,
\end{equation}
where $\phi: PC \rightarrow PC_T$ is a bijection.

%\duygu{This is my attempt to describe the two passes, please check if this makes sense.}
We compute these distances between point sets sampled both on the source (using the DMSO) and target models. Moreover, computing the above losses on point sets sampled on source and target models further helps for robustness to different mesh densities. In practice, for each $(S,T)$ source-target model pair, we also pass a point cloud sampled on $S$ together with $T$ through the decoder offset in a second pass to help the network cope with sparse meshes. Specifically, given a point set sampled on $S$, we predict per-point offsets and compute the above Chamfer and Earth Mover's losses between the resulting deformed point cloud and $T$. We denote these two losses as $L_{\texttt{CD}}^{\texttt{Points}}$ and $L_{\texttt{EMD}}^{\texttt{Points}}$. During testing, this second pass is not necessary and we only predict per-vertex offsets for $S$.

We note that we train our model with synthetic data where we always have access to 3D models. Thus, even if the target is a 2D image, we use the corresponding 3D model to compute the point cloud shape loss. During testing, however, we do not need access to any 3D target models, since the global shape features required for offset prediction are extracted from the 2D image only.

\paragraph{Symmetry Loss.} Many man-made models exhibit global reflection symmetry and our goal is to preserve this during deformation. However, the mesh topology itself does not always guarantee to be symmetric, i.e., a symmetric chair does not always have symmetric vertices. Therefore, we propose to preserve shape symmetry by sampling a point cloud, $M(PC)$, on the mirrored deformed output and measure the point cloud shape loss with this mirrored point cloud as
\begin{align}
L_{\texttt{sym}}(PC, PC_T) &= L_{CD}(M(PC), PC_T) \nonumber \\
&+ L_{EMD}(M(PC), PC_T).
\end{align}
We note that we assume the reflection symmetry plane of a source model to be known. In our experiments, we use 3D models from ShapeNet~\cite{chang2015shapenet} which are already aligned such that the reflection plane coincides with the $xz-$ plane.

\paragraph{Mesh Laplacian Loss.} To preserve the local geometric details in the source mesh and enforce smooth deformation across the mesh surface, we desire the Laplacian coordinates of the deformed mesh to be the same as the original source mesh. We define this loss as 
\vspace{-5pt}
\begin{equation}
L_{\texttt{lap}} = \sum_i|| Lap(S) - Lap(S') ||_2.\,
\end{equation}
where $Lap$ is the mesh Laplacian operator, $S$ and $S'$ are the original and deformed meshes respectively.

\begin{figure}
	\begin{center}
		\begin{tabular}{ccc}
			\includegraphics[height=0.11\textwidth]{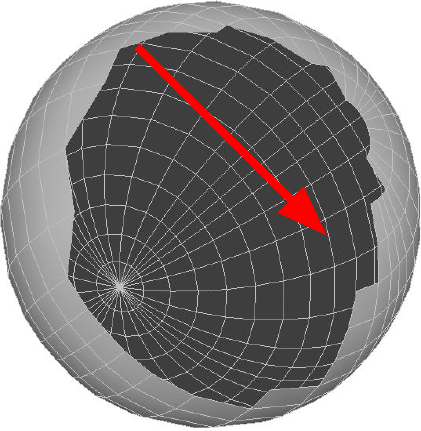} &
			\includegraphics[height=0.11\textwidth]{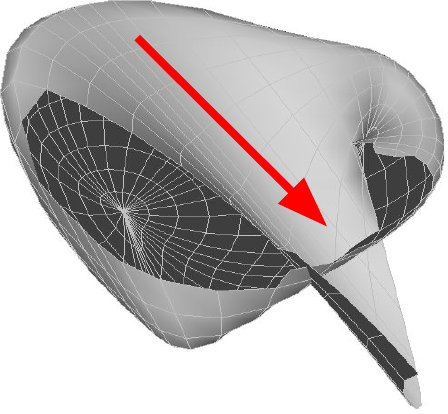} &
			\includegraphics[height=0.10\textwidth]{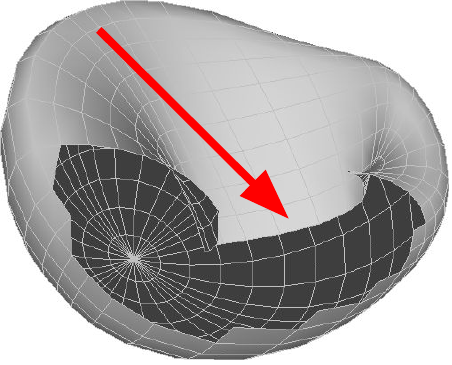}
			\\
			(a)&(b)&(c)
			
		\end{tabular}
	\end{center}
	\caption{Self intersection. The red arrow is the deformation handle. (a) Original Mesh. (b) Deformation with self-intersection. (c) Plausible deformation.}
	\vspace{-6pt}
	\label{fig:localpermutation}
\end{figure}

\vspace{-5pt}
\paragraph{Local Permutation Invariant Loss.} Most traditional deformation methods (such as FFD) are prone to suffer from possible self-intersections that can occur during deformation (see Figure~\ref{fig:localpermutation}). To prevent such self-intersections, we present a novel \emph{local permutation invariant loss}. Specifically, given a point $p$ and a neighboring point at a distance $\delta$ to $p$, we would like to preserve the distance between these two neighboring points after deformation as well. Thus, we define
\begin{equation}
L_{\texttt{LPI}} = -\min{(F(V + \delta) - F(V), \mathbf{0})}.\,
\end{equation}
where $\delta$ is a vector with a small magnitude and $\mathbf{0} = (0,0,0)$. In our experiments we define $\delta \in \{(\epsilon, 0, 0), (0, \epsilon, 0), (0, 0, \epsilon)\} $ where $\epsilon=0.05$. The intuition behind of this is to preserve the local ordering of points in the source. We observe that the local permutation invariant loss helps to achieve smooth deformation across 3D space.
%For example, if point A is on the left of point B, in the deformed shape, A' should still be on the left of B'. 
%Figure~\ref{fig:loss} illustrates a pipeline for computing different losses. 
Given all the losses defined above, we train 3DN with a combined loss of
\vspace{-5pt}
\begin{align}
L = \omega_{L_1} L_{\texttt{CD}}^{\texttt{Mesh}} + \omega_{L_2} L_{\texttt{EMD}}^{\texttt{Mesh}} + \omega_{L_3} L_{\texttt{CD}}^{\texttt{Points}} + \omega_{L_4} L_{\texttt{EMD}}^{\texttt{Points}} +  \nonumber\\
\omega_{L_5} L_{\texttt{sym}} + \omega_{L_6} L_{\texttt{lap}} + \omega_{L_7} L_{\texttt{LPI}},
\end{align}
where $\omega_{L_1}, \omega_{L_2}, \omega_{L_3}, \omega_{L_4}, \omega_{L_5}, \omega_{L_6}, \omega_{L_7}$ denote the relative weighting of the losses. 

%\begin{figure}

%% file: evaluation.tex
%-------------------------------------------------------------------------
\begin{figure*}
	\centering
	\newcolumntype{C}{>{\centering\arraybackslash}p{2.6em}}
	\newcolumntype{D}{>{\centering\arraybackslash}p{1.2em}}
	\newcolumntype{E}{ >{\centering\arraybackslash}m{0.4em}}
	\renewcommand{\arraystretch}{0.1}% Tighter
	\centering
	\resizebox{\textwidth}{!}{%resizing the whole table
	\begin{tabular}{cccccccc}

\includegraphics[height=0.07\textwidth]{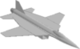} &
\includegraphics[height=0.07\textwidth]{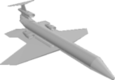} &
\includegraphics[height=0.07\textwidth]{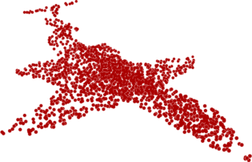}&
\includegraphics[height=0.07\textwidth]{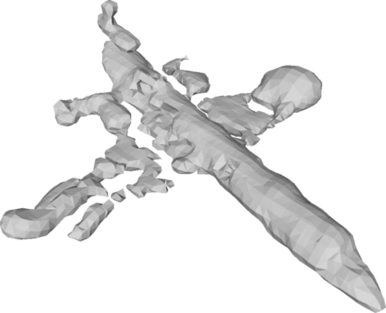}&
\includegraphics[height=0.07\textwidth]{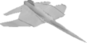}&
\includegraphics[height=0.07\textwidth]{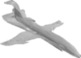} &
\includegraphics[height=0.07\textwidth]{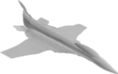} &
\\

\includegraphics[height=0.07\textwidth]{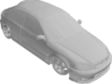} &
\includegraphics[height=0.07\textwidth]{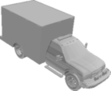} &
\includegraphics[height=0.07\textwidth]{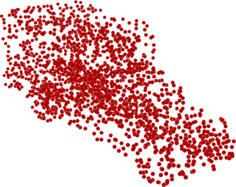}&
\includegraphics[height=0.07\textwidth]{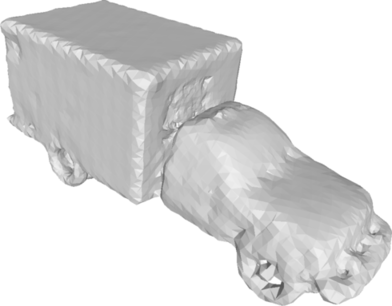}&
\includegraphics[height=0.07\textwidth]{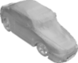}&
\includegraphics[height=0.07\textwidth]{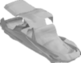} &
\includegraphics[height=0.07\textwidth]{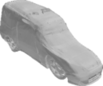} 
\\

\includegraphics[height=0.13\textwidth]{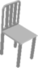} &
\includegraphics[height=0.1\textwidth]{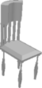} &
\includegraphics[height=0.1\textwidth]{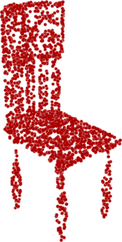}&
\includegraphics[height=0.1\textwidth]{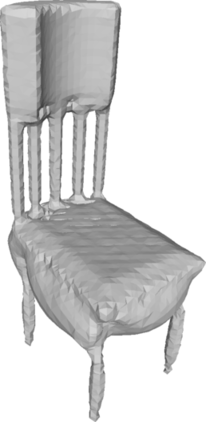}&
\includegraphics[height=0.1\textwidth]{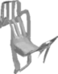}&
\includegraphics[height=0.1\textwidth]{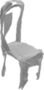} &
\includegraphics[height=0.1\textwidth]{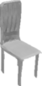} 
\\

\includegraphics[height=0.07\textwidth]{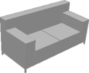} &
\includegraphics[height=0.07\textwidth]{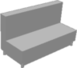} &
\includegraphics[height=0.07\textwidth]{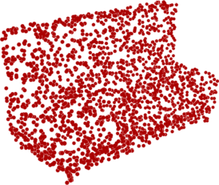}&
\includegraphics[height=0.07\textwidth]{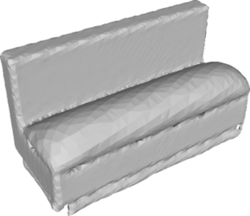}&
\includegraphics[height=0.07\textwidth]{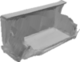}&
\includegraphics[height=0.07\textwidth]{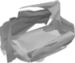} &
\includegraphics[height=0.07\textwidth]{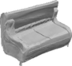} 
\\

\includegraphics[height=0.07\textwidth]{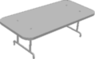} &
\includegraphics[height=0.07\textwidth]{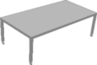} &
\includegraphics[height=0.07\textwidth]{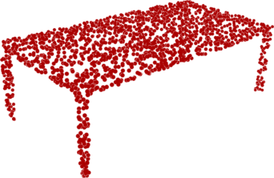}&
\includegraphics[height=0.07\textwidth]{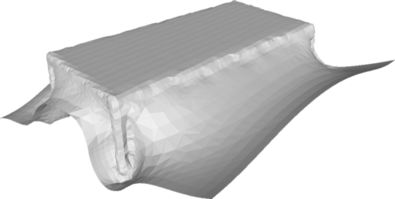}&
\includegraphics[height=0.07\textwidth]{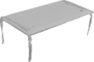}&
\includegraphics[height=0.07\textwidth]{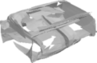} &
\includegraphics[height=0.07\textwidth]{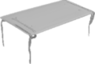} 
\\
\includegraphics[height=0.07\textwidth]{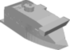} &
\includegraphics[height=0.07\textwidth]{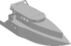} &
\includegraphics[height=0.07\textwidth]{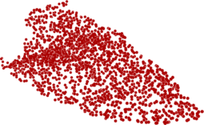}&
\includegraphics[height=0.07\textwidth]{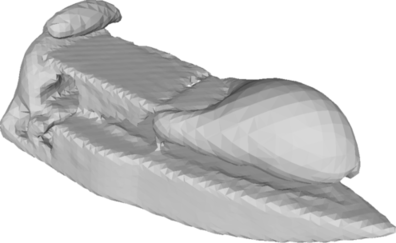}&
\includegraphics[height=0.07\textwidth]{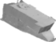}&
\includegraphics[height=0.07\textwidth]{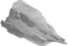} &
\includegraphics[height=0.07\textwidth]{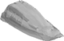} 
\\
(a) Source Template&(b) Target Mesh&(c) Target Point Cloud&(d)Poisson&(e)FFD&(f)AtlasNet &(g) Ours
	\end{tabular}}
	\caption{Given a source (a) and a target (b) model from the ShapeNet dataset, we show the deformed meshes obtained by our method (g). We also show Poisson surface reconstruction (d) from a set of points sampled on the target (c). We also show comparisons to previous methods of Jack et al. (e) and AtlasNet (f).}
	\label{fig:3d}
\end{figure*}

\vspace{-4pt}
\section{Experiments}
\vspace{-4pt}
In this section, we perform qualitative and quantitative comparisons on shape reconstruction from 3D target models (Section~\ref{sec:3d}) as well as single-view reconstruction (Section~\ref{sec:2d}). We also conduct ablation studies of our method to demonstrate the effectiveness of the offset decoder architecture and the different loss functions employed. Finally, we provide several applications to demonstrate the flexibility of our method. More qualitative results and implementation details can be found in supplementary material.
\vspace{-8pt}
\paragraph{Dataset.} In our experiments, we use the ShapeNet Core dataset~\cite{chang2015shapenet} which includes 13 shape categories and an official traning/testing split. We use the same template set of models as in \cite{jack2018learning} for potential source meshes. There are 30 shapes for each category in this template set. When training the 2D image-based target model, we use the rendered views provided by Choy et al.~\cite{choy20163d}. We note that we train a single network across all categories.
\vspace{-8pt}
\paragraph{Template Selection.} In order to sample source and target model pairs for 3DN, we train a PointNet based auto-encoder to learn an embedding of the 3D shapes. Specifically, we represent each 3D shape as a uniformly sampled set of points. The encoder encodes the points as a feature vector and the decoder predicts the point positions from this feature vector (please refer to the supplementary material for details). Given the embedding composed of the features extracted by the encoder, for each target model candidate, we choose the nearest neighbor in this embedding as the source model. Source models are chosen from the aforementioned template set. No class label information is required during this procedure, however, the nearest neighbors are queried within the same category. When given a target 2D image for testing, if no desired source model is given, we use the point set generation network, PSGN~\cite{fan2017point}, to generate an initial point cloud, and use its nearest neighbor in our embedding as the source model.
\vspace{-8pt}
\paragraph{Evaluation Metrics.} Given a source and target model pair $(S,T)$, we utilize three metrics in our quantitative evaluations to compare the deformation output $S'$ and the target $T$: 1) Chamfer Distance (CD) between the point clouds sampled on $S'$ and $T$, 2) Earth Mover's Distance (EMD) between the point clouds sampled on $S'$ and $T$, 3) Intersection over Union (IoU) between the solid voxelizations of $S'$ and $T$. We normalize the outputs of our method and previous work into a unit cube before computing these metrics. %Since different methods are using different coordinate system, we normalize the output and performed the comparison. 
We also evaluate the visual plausibility of our results by providing a large set of qualitative examples.
\vspace{-8pt}
\paragraph{Comparison} We compare our approach with state-of-the-art reconstruction methods. Specifically, we compare to three categories of methods: 1) learning-based surface generation, 2) learning-based deformation prediction, and 3) traditional surface reconstruction methods. We would like to note that we are solving a fundamentally different problem than surface generation methods. Even though, having a source mesh to start with might seem advantageous, our problem at hand is not easier since our goal is not only to generate a mesh similar to the target but also preserve certain properties of the source. Furthermore, our source meshes are obtained from a fixed set of templates which contain only 30 models per category. 
\begin{table*}
	\begin{center}
		\setlength{\tabcolsep}{0.1em}
		\newcolumntype{C}{>{\centering\arraybackslash}p{2.5em}}
			\begin{tabular}{C|c|CCCCCCCCCCCCC|C}
				\Xhline{3\arrayrulewidth}
				& \small & plane& bench& box& car& chair& display& lamp& speaker& rifle& sofa& table&phone&boat&Mean\\
				\hline
				\multirow{4}{*}{EMD}
				&AtlasNet&3.46 & 3.18 & 4.20 & 2.84 & 3.47 & 3.97 & 3.79 & 3.83 & 2.44 & 3.19 & 3.76 & 3.87 & 2.99 & 3.46 \\
				&FFD&1.88 & 2.02 & 2.50 & 2.11 & 2.13 & 2.69 & 2.42 & 3.06 & 1.55 & 2.44 & 2.44 & 1.88 & 2.00 & \textbf{2.24} \\
				&Ours&0.79 &1.98 &3.57 &1.24 &1.12 &3.08 &3.44 &3.40 &1.79 &2.06 &1.34 &3.27 &2.27 &2.26 \\
				\hline
				\multirow{4}{*}{CD}
				&AtlasNet&2.16 & 2.91 & 6.62 & 3.97 & 3.65 & 3.65 & 4.48 & 6.29 & 0.98 & 4.34 & 6.01 & 2.44 & 2.73 & 3.86 \\
				&FFD&3.22 & 4.53 & 6.94 & 4.45 & 4.99 & 5.98 & 8.72 & 11.97 & 1.97 & 6.29 & 6.89 & 3.61 & 4.41 & 5.69 \\
				&Ours&0.38 &2.40 &5.26 &0.90 &0.82 &5.59 &8.74 &9.27 &1.52 &2.55 &0.97 &2.66 &2.77 &\textbf{3.37} \\
				\hline
				\multirow{4}{*}{IoU}&AtlasNet& 56.9 & 53.3 & 31.3 & 44.0 & 47.9 & 48.0 & 41.6 & 33.2&63.4&44.7&43.8&58.7&50.9&46.7\\
				&FFD& 29.0 &42.3 &28.4 &21.1 &42.2 &27.9 &38.9 &52.5 &31.9 &34.7 &43.3 &22.9 &47.7 & 35.6\\
				&Ours& 71.0 &40.7 &43.6 &75.8 &66.3 &40.4 &25.1 &49.2 &40.0 &60.6 &57.9 &50.1 &42.6 &\textbf{51.1}\\
				
				\Xhline{3\arrayrulewidth}
			\end{tabular}
	\end{center}
	\caption{Point cloud reconstruction results on ShapeNet core dataset. Metrics are mean Chamfer distance ($\times 0.001$, CD) on points, Earth Mover's distance ($\times 100$, EMD) on points and Intersection over Union (\%, IoU) on solid voxelized grids. For both CD and EMD, the lower the better. For IoU, the higher the better. }
	\label{table:3d}
	\vspace{-10pt}
\end{table*}

\subsection{Shape Reconstruction from Point Cloud}
\label{sec:3d}
For this experiment, we define each 3D model in the testing split as target and identify a source model in the testing split based on the autoencoder embedding described above.
3DN computes per-vertex displacements to deform the source and keeps the source mesh topology fixed. We evaluate the quality of this mesh with alternative meshing techniques. %Specifically, given a source and target model pair, we uniformly sample a set of points on the source and predict per-point displacements for each point to match the target. Given the deformed set of points, we reconstruct a 3D mesh using Poisson surface reconstruction.
Specifically, given a set of points sampled on the desired target model, we reconstruct a 3D mesh using Poisson surface reconstruction. As shown in Figure~\ref{fig:3d}, this comparison demonstrates that even with a ground truth set of points, generating a mesh that preserves sharp features is not trivial. Instead, our method utilizes the source mesh connectivity to output a plausible mesh. Furthermore, we apply the learning-based surface generation technique of AtlasNet~\cite{groueix2018} on the uniformly sampled points on the target model. Thus, we expect AtlasNet only to perform surface generation without any deformation. We also compare to the method of Jack et al.~\cite{jack2018learning} (FFD) which introduces a learning based method to apply free form deformation to a given template model to match an input image. This network consists of a module which predicts FFD parameters based on the features extracted from the input image. We retrain this module such that it uses the features extracted from the points sampled on the 3D target model. As shown in Figure~\ref{fig:3d}, the deformed meshes generated by our method are higher quality than the previous methods. We also report quantitative numbers in Table~\ref{table:3d}. While AtlastNet achieves lower error based on Chamfer Distance, we observe certain artifacts such as holes and disconnected surfaces in their results. We also observe that our deformation results are smoother than FFD. 
\vspace{-15pt} 
\subsection{Single-view Reconstruction}
\label{sec:2d}
%Pixel2mesh: https://arxiv.org/pdf/1804.01654.pdf
%ffd with template: https://arxiv.org/pdf/1803.10932.pdf
%AtlasNet: https://arxiv.org/pdf/1802.05384.pdf

\begin{table*}
	\begin{center}
		\setlength{\tabcolsep}{0.1em}
		\newcolumntype{C}{>{\centering\arraybackslash}p{2.5em}}
		\begin{tabular}{C|c|CCCCCCCCCCCCC|C}
			\Xhline{3\arrayrulewidth}
			& \small & plane& bench& box& car& chair& display& lamp& speaker& rifle& sofa& table&phone&boat&Mean\\
			\hline
			\multirow{4}{*}{EMD}  &AtlasNet&3.39 & 3.22 & 3.36 & 3.72 & 3.86 & 3.12 & 5.29 & 3.75 & 3.35 & 3.14 & 3.98 & 3.19 & 4.39 & 3.67  \\
			&Pxel2mesh&2.98 & 2.58 & 3.44 & 3.43 & 3.52 & 2.92 & 5.15 & 3.56 & 3.04 & 2.70 & 3.52 & 2.66 & 3.94 & \textbf{3.34} \\ 
			&FFD&2.63 & 3.96 & 4.87 & 2.98 & 3.38 & 4.88 & 7.19 & 5.04 & 3.58 & 3.70 & 3.56 & 4.11 & 3.86 & 4.13  \\
			&Ours&3.30 &2.98 &3.21 &3.28 &4.45 &3.91 &3.99 &4.47 &2.78 &3.31 &3.94 &2.70 &3.92 &3.56\\
			\hline
			\multirow{4}{*}{CD} &AtlasNet&5.98 & 6.98 & 13.76 & 17.04 & 13.21 & 7.18 & 38.21 & 15.96 & 4.59 & 8.29 & 18.08 & 6.35 & 15.85 & 13.19 \\ 
			&Pixel2mesh&6.10 & 6.20 & 12.11 & 13.45 & 11.13 & 6.39 & 31.41 & 14.52 & 4.51 & 6.54 & 15.61 & 6.04 & 12.66 & 11.28  \\
			&FFD&3.41 & 13.73 & 29.23 & 5.35 & 7.75 & 24.03 & 45.86 & 27.57 & 6.45 & 11.89 & 13.74 & 16.93 & 11.31 & 16.71  \\
			&Ours&6.75 &7.96 &8.34 &7.09 &17.53 &8.35 &12.79 &17.28 &3.26 &8.27 &14.05 &5.18 &10.20 &\textbf{9.77}\\
			\hline
			\multirow{4}{*}{IoU} &AtlasNet&39.2&34.2&20.7& 22.0&25.7&36.4&21.3&23.2&45.3&27.9&23.3&42.5&28.1&30.0\\
			&Pixel2mesh& 51.5&40.7&43.4&50.1&40.2&55.9&29.1&52.3&50.9&60.0&31.2&69.4&40.1&47.3\\
			&FFD&30.3 &44.8 &30.1 &22.1 &38.7 &31.6 &35.0 &52.5 &29.9 &34.7 &45.3 &22.0 &50.8 &36.7\\
			&Ours&54.3 &39.8 &49.4 &59.4 &34.4 &47.2 &35.4 &45.3 &57.6 &60.7 &31.3 &71.4 &46.4 &\textbf{48.7}\\
			\Xhline{3\arrayrulewidth}
		\end{tabular}
	\end{center}
	\caption{Quantitative comparison on ShapeNet rendered images. Metrics are CD ($\times 0.001$), EMD ($\times 100$) and IoU (\%).}
	\label{table:2d}
	\vspace{-10pt}
	%\vspace{-3mm}
\end{table*}
We also compare our method to recent state-of-the-art single view image based reconstruction methods including Pixel2Mesh~\cite{wang2018pixel2mesh}, AtlasNet~\cite{groueix2018} and FFD~\cite{jack2018learning}. Specifically, we choose a target rendered image from the testing split and input to the previous methods. For our method, in addition to this target image, we also provide a source model selected from the template set. We note that the scope of our work is not single-view reconstruction, thus the comparison with Pixel2Mesh and AtlasNet is not entirely fair. However, both quantitative (see Table~\ref{table:2d}) and qualitative (Figure~\ref{fig:2d}) results still provide useful insights. Though the rendered output of AtlasNet and Pixel2Mesh in Figure~\ref{fig:2d} are visually plausible, self-intersections and disconnected surfaces often exist in their results. Figure~\ref{fig:Closeup} illustrates this by rendering the output meshes in wireframe mode. Furthermore, as shown in Figure~\ref{fig:Closeup}, while surface generation methods struggle to capture shape details such as chair handles and car wheels, our method preserves these details that reside in the source mesh.
%\vspace{-10pt}

\begin{figure}
	\hspace*{-15pt}
	\centering
	\newcolumntype{C}{>{\centering\arraybackslash}p{2.3em}}
	\newcolumntype{D}{>{\centering\arraybackslash}p{3.5em}}
	\newcolumntype{E}{>{\centering\arraybackslash}p{.25em}}
	\renewcommand{\arraystretch}{0.1}% Tighter
%	\begin{tabular}{ECCDDcccccDc}
	\begin{tabular}{CCCCCCC}	

\includegraphics[width=0.05\textwidth]{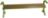} &
\includegraphics[width=0.05\textwidth]{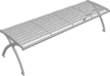} &
\includegraphics[width=0.05\textwidth]{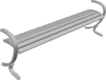} &
\includegraphics[width=0.05\textwidth]{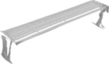} &
\includegraphics[width=0.05\textwidth]{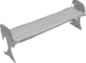} &
\includegraphics[width=0.05\textwidth]{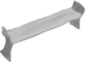} &
\includegraphics[width=0.05\textwidth]{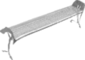} 
\\

\includegraphics[width=0.05\textwidth]{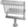} &
\includegraphics[width=0.05\textwidth]{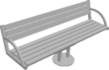} &
\includegraphics[width=0.05\textwidth]{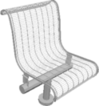}&
\includegraphics[width=0.05\textwidth]{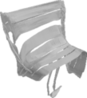}&
\includegraphics[width=0.05\textwidth]{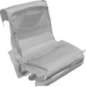} &
\includegraphics[width=0.05\textwidth]{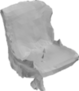} &
\includegraphics[width=0.05\textwidth]{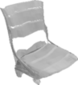} 
\\

\includegraphics[width=0.065\textwidth]{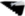} &
\includegraphics[width=0.065\textwidth]{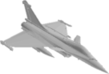} &
\includegraphics[width=0.065\textwidth]{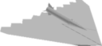}&
\includegraphics[width=0.065\textwidth]{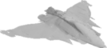}&
\includegraphics[width=0.065\textwidth]{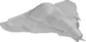} &
\includegraphics[width=0.065\textwidth]{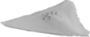} &
\includegraphics[width=0.065\textwidth]{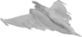} 
\\

\includegraphics[height=0.065\textwidth]{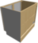} &
\includegraphics[height=0.065\textwidth]{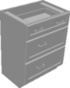} &
\includegraphics[height=0.065\textwidth]{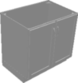}&
\includegraphics[height=0.065\textwidth]{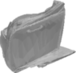}&
\includegraphics[height=0.065\textwidth]{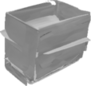} &
\includegraphics[height=0.065\textwidth]{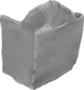} &
\includegraphics[height=0.065\textwidth]{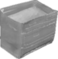} 
\\

\includegraphics[height=0.065\textwidth]{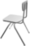} &
\includegraphics[height=0.065\textwidth]{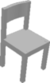} &
\includegraphics[height=0.065\textwidth]{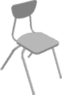}&
\includegraphics[height=0.065\textwidth]{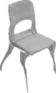}&
\includegraphics[height=0.065\textwidth]{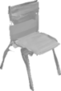} &
\includegraphics[height=0.065\textwidth]{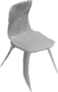} &
\includegraphics[height=0.065\textwidth]{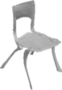} 
\\

\includegraphics[height=0.065\textwidth]{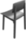} &
\includegraphics[height=0.065\textwidth]{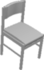} &
\includegraphics[height=0.065\textwidth]{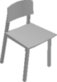}&
\includegraphics[height=0.065\textwidth]{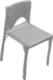}&
\includegraphics[height=0.065\textwidth]{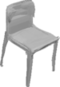} &
\includegraphics[height=0.065\textwidth]{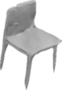} &
\includegraphics[height=0.065\textwidth]{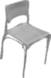} 
\\

\includegraphics[height=0.065\textwidth]{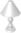} &
\includegraphics[height=0.065\textwidth]{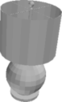} &
\includegraphics[height=0.065\textwidth]{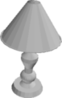}&
\includegraphics[height=0.065\textwidth]{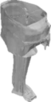}&
\includegraphics[height=0.065\textwidth]{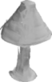} &
\includegraphics[height=0.065\textwidth]{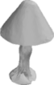} &
\includegraphics[height=0.065\textwidth]{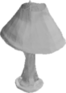} 
\\

\includegraphics[height=0.065\textwidth]{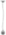} &
\includegraphics[height=0.065\textwidth]{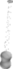} &
\includegraphics[height=0.065\textwidth]{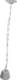}&
\includegraphics[height=0.065\textwidth]{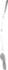}&
\includegraphics[height=0.065\textwidth]{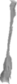} &
\includegraphics[height=0.065\textwidth]{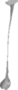} &
\includegraphics[height=0.065\textwidth]{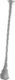} 
\\

\includegraphics[height=0.065\textwidth]{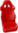} &
\includegraphics[height=0.065\textwidth]{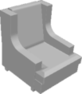} &
\includegraphics[height=0.065\textwidth]{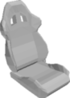}&
\includegraphics[height=0.065\textwidth]{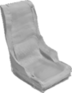}&
\includegraphics[height=0.065\textwidth]{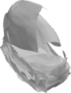} &
\includegraphics[height=0.065\textwidth]{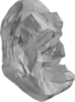} &
\includegraphics[height=0.065\textwidth]{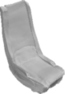} 
\\

\includegraphics[height=0.065\textwidth]{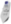} &
\includegraphics[width=0.065\textwidth]{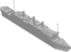} &
\includegraphics[width=0.065\textwidth]{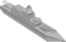}&
\includegraphics[width=0.065\textwidth]{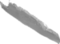}&
\includegraphics[width=0.065\textwidth]{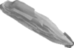} &
\includegraphics[width=0.065\textwidth]{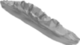} &
\includegraphics[width=0.065\textwidth]{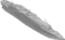} 
\\

\includegraphics[height=0.055\textwidth]{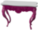}  &
\includegraphics[height=0.065\textwidth]{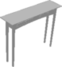}  &
\includegraphics[height=0.065\textwidth]{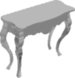} &
\includegraphics[height=0.065\textwidth]{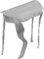} &
\includegraphics[height=0.065\textwidth]{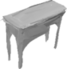}  &
\includegraphics[height=0.065\textwidth]{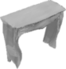}  &
\includegraphics[height=0.065\textwidth]{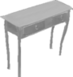} 
\\
\small{Target}&\small{Source}&\small{GT}&\small{FFD}&\small{AtlasNet}&\small{P2M}&\small{3DN}
	\end{tabular}
	\caption{Given a target image and a source, we show deformation results of FFD, AtlasNet, Pixel2Mesh (P2M), and 3DN. We also show the ground truth target model (GT).}
	\vspace{-17pt}
	%\vspace{-3mm}
	\label{fig:2d}
\end{figure}

\begin{figure}
	\centering
	\renewcommand{\arraystretch}{0.1}% Tighter
	\newcolumntype{C}{>{\centering\arraybackslash}p{3em}}
	\hspace*{-20pt}
	\begin{tabular}{CCCCCC}
		\includegraphics[width=0.08\textwidth]{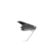} &
		\includegraphics[width=0.08\textwidth]{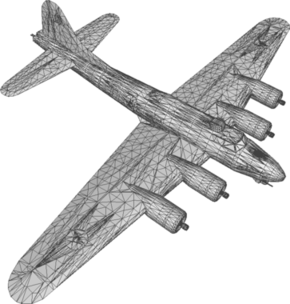} &
		\includegraphics[width=0.08\textwidth]{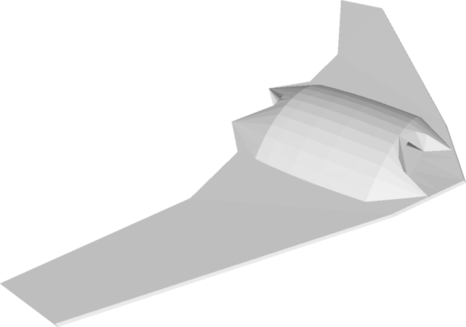} &
		\includegraphics[width=0.08\textwidth]{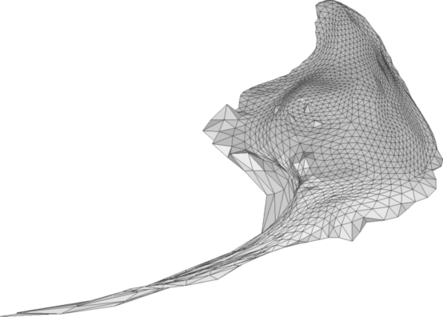} &
		\includegraphics[width=0.08\textwidth]{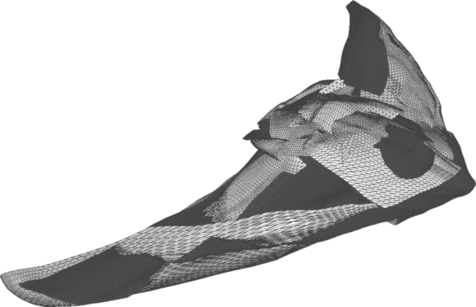} &
		\includegraphics[width=0.08\textwidth]{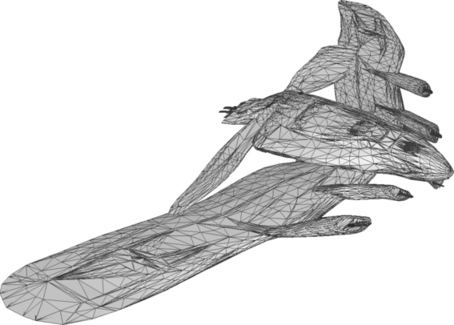} 
		\\
		\includegraphics[width=0.08\textwidth]{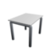} &
		\includegraphics[width=0.08\textwidth]{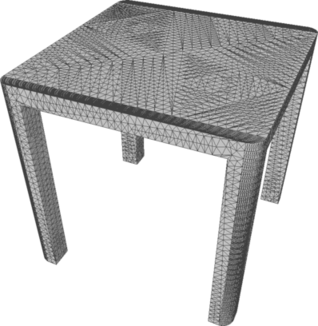} &
		\includegraphics[width=0.08\textwidth]{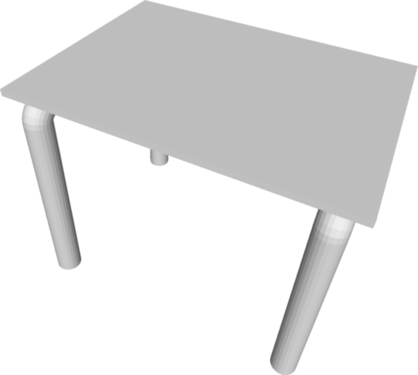} &
		\includegraphics[width=0.08\textwidth]{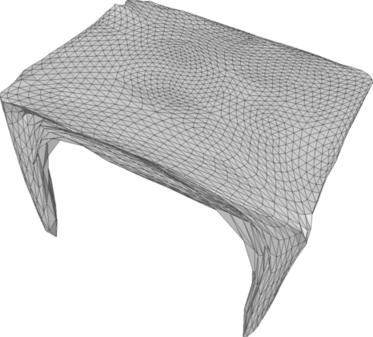} &
		\includegraphics[width=0.08\textwidth]{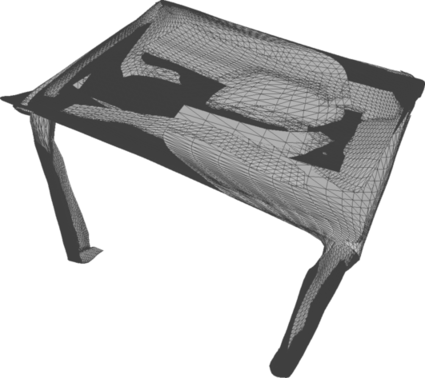} &
		\includegraphics[width=0.08\textwidth]{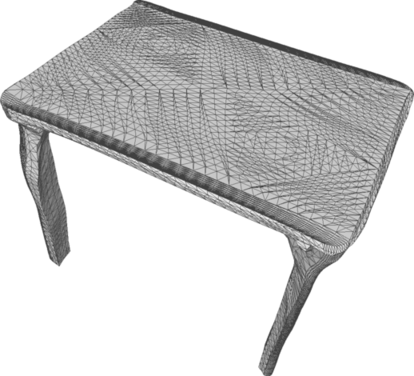} 
		\\
		\includegraphics[width=0.08\textwidth]{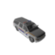} &
		\includegraphics[width=0.08\textwidth]{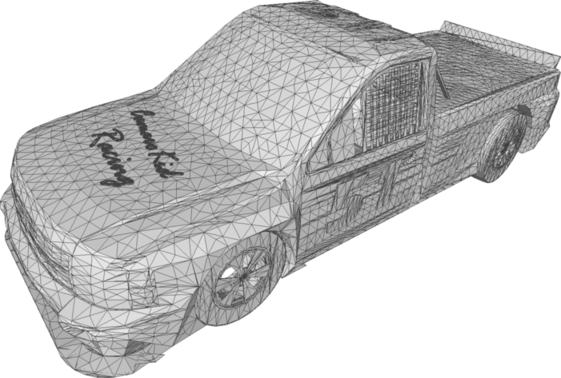} &
		\includegraphics[width=0.08\textwidth]{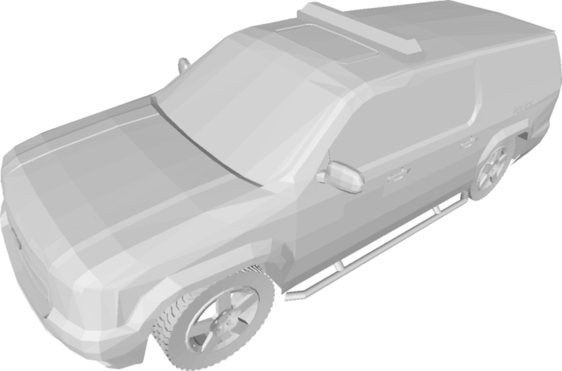} &
		\includegraphics[width=0.08\textwidth]{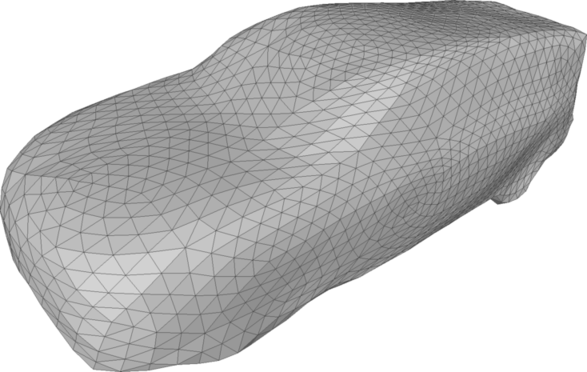} &
		\includegraphics[width=0.08\textwidth]{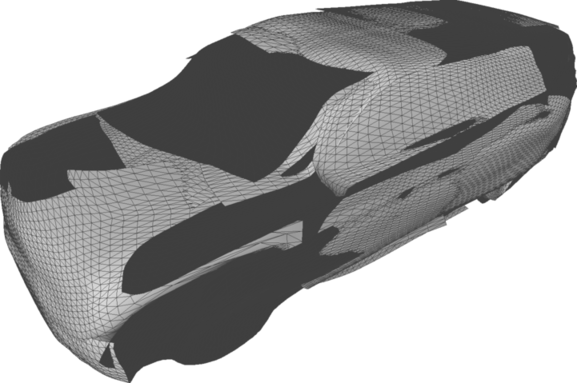} &
		\includegraphics[width=0.08\textwidth]{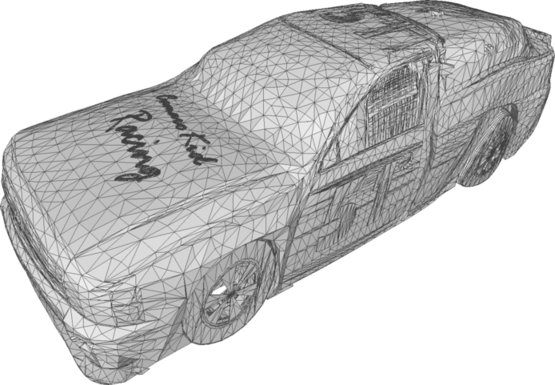} 
		\\
		\includegraphics[width=0.08\textwidth]{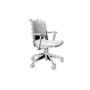} &
		\includegraphics[width=0.08\textwidth]{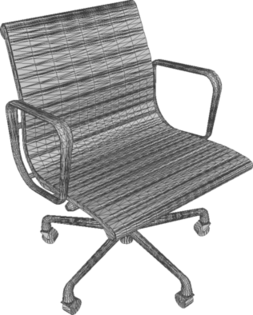} &
		\includegraphics[width=0.08\textwidth]{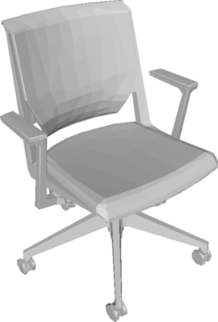} &
		\includegraphics[width=0.065\textwidth]{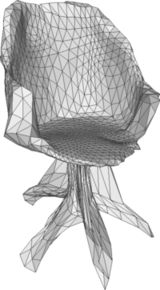} &
		\includegraphics[width=0.09\textwidth]{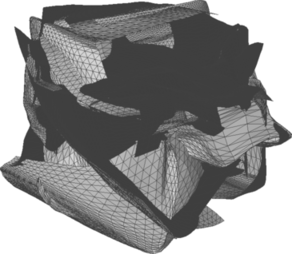} &
		\includegraphics[width=0.07\textwidth]{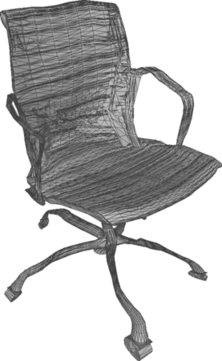} 
		\\
		\centering \small{Target}&\small{Source} & \small{GT} & \small{P2M} & \small{AtlasNet} & \small{3DN}
	\end{tabular}
	\caption{For a given target image and source model, we show ground truth model and results of Pixel2Mesh (P2M), AtlasNet, and our method (3DN) rendered in wire-frame mode to better judge the quality of the meshes. Please zoom into the PDF for details.}
	%\vspace{-10pt}
	\label{fig:Closeup}
\end{figure}
\begin{figure}
	\centering
	\renewcommand{\arraystretch}{0.1}% Tighter
	\newcolumntype{C}{>{\centering\arraybackslash}p{3em}}
	\hspace*{-20pt}
	\begin{tabular}{CCCCCC}
		\includegraphics[width=0.08\textwidth]{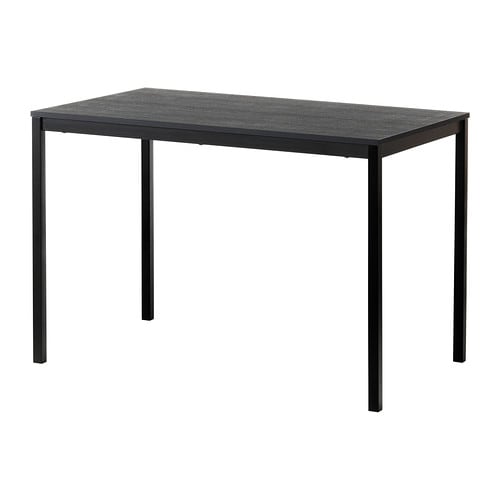} &
		\includegraphics[width=0.08\textwidth]{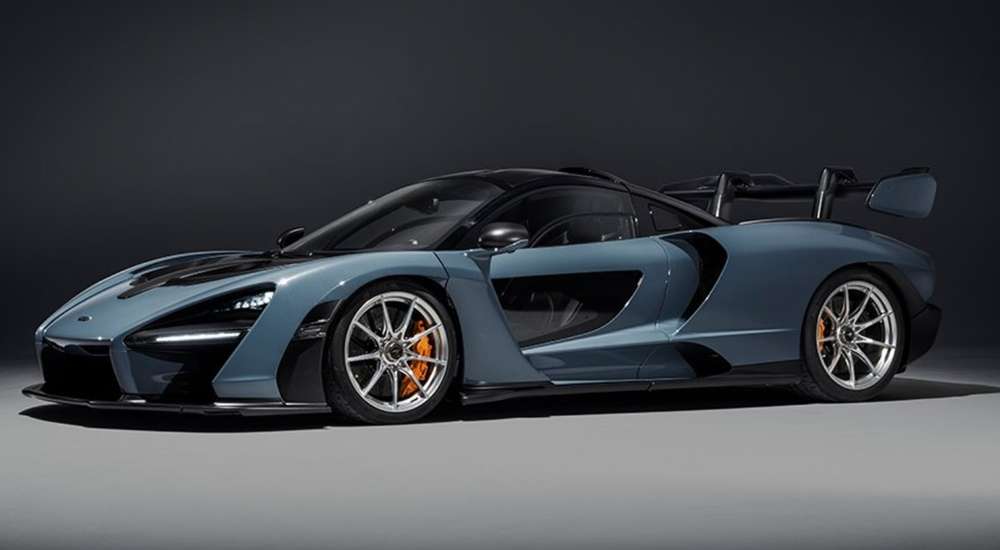} &
		\includegraphics[width=0.08\textwidth]{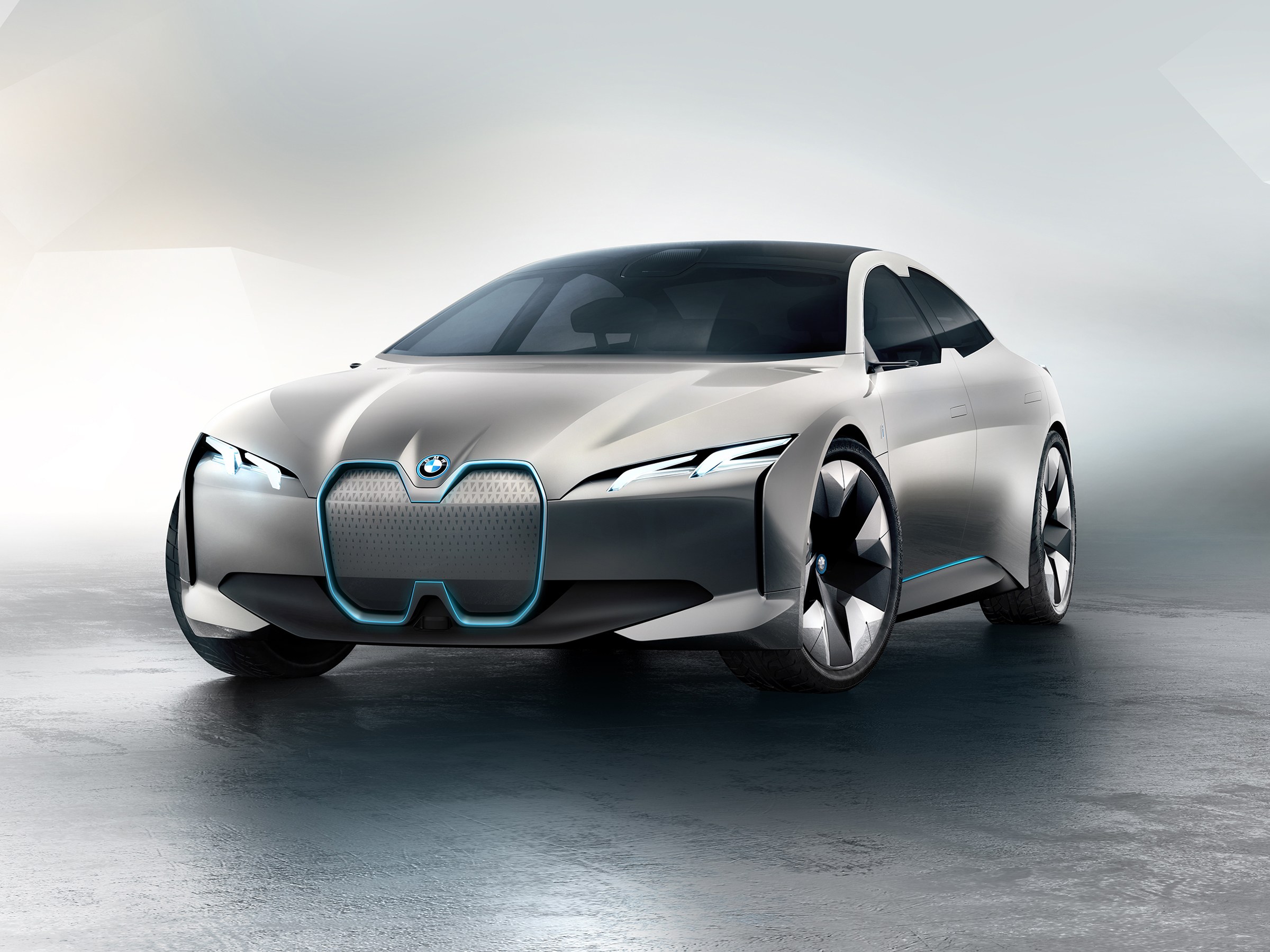} &
		\includegraphics[width=0.08\textwidth]{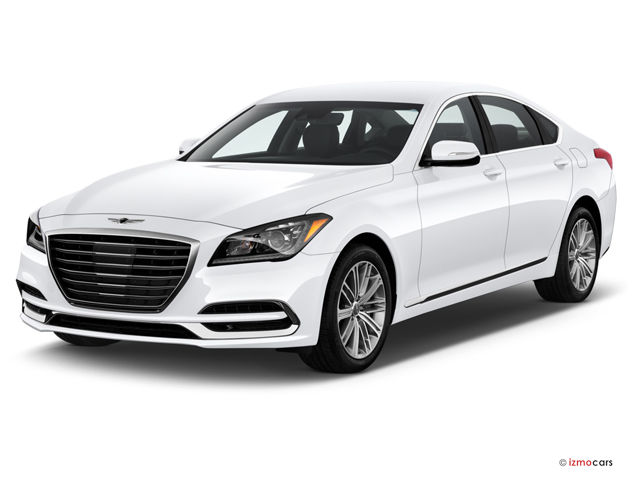} &
		\includegraphics[width=0.08\textwidth]{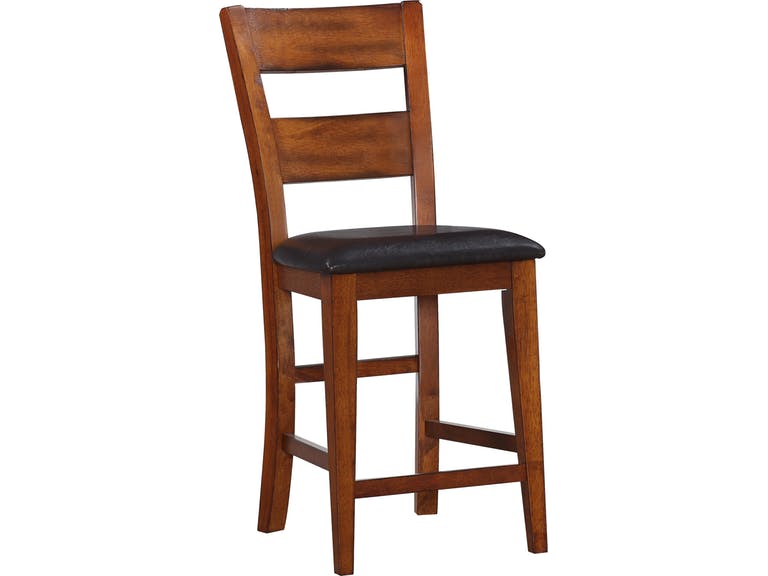} &
		\includegraphics[width=0.08\textwidth]{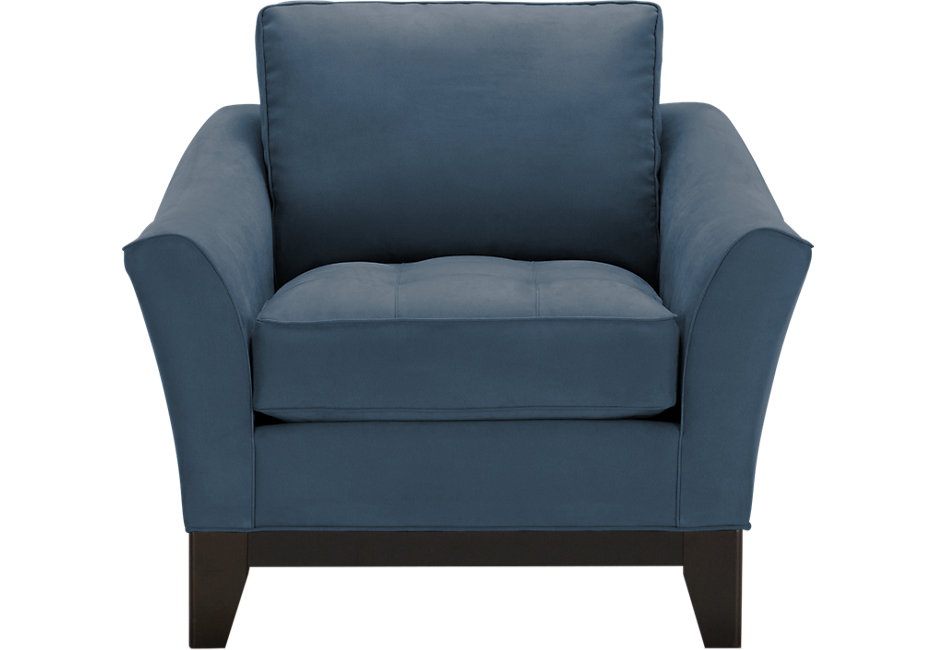}
		\\
		\includegraphics[width=0.08\textwidth]{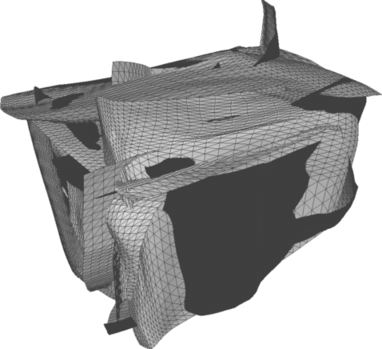}&
		\includegraphics[width=0.08\textwidth]{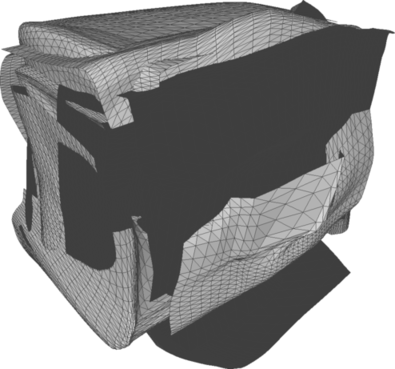} &
		\includegraphics[width=0.08\textwidth]{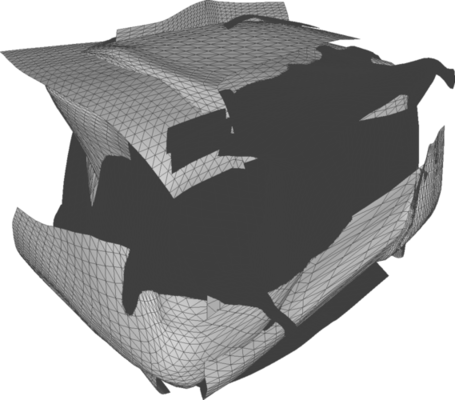} &
		\includegraphics[width=0.08\textwidth]{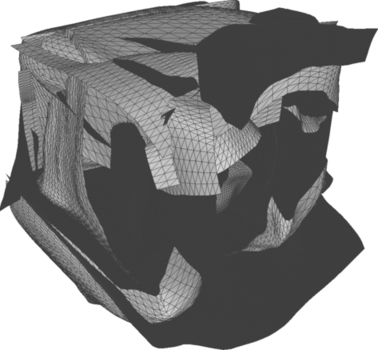} &
		\includegraphics[width=0.08\textwidth]{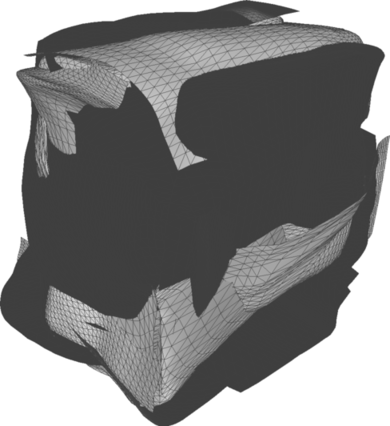} &
		\includegraphics[width=0.08\textwidth]{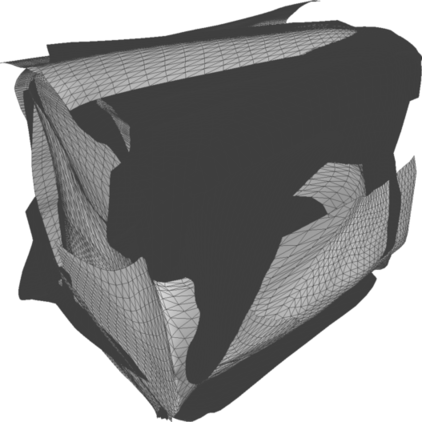} 
		\\
		\includegraphics[width=0.065\textwidth]{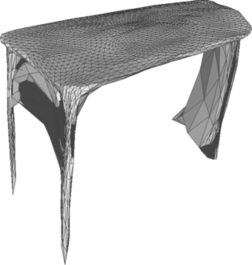}&
		\includegraphics[width=0.08\textwidth]{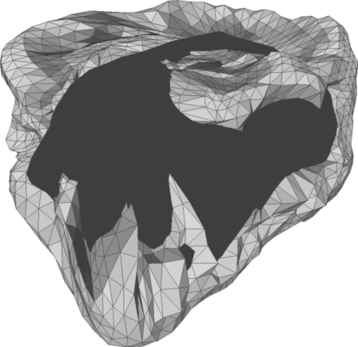} &
		\includegraphics[width=0.08\textwidth]{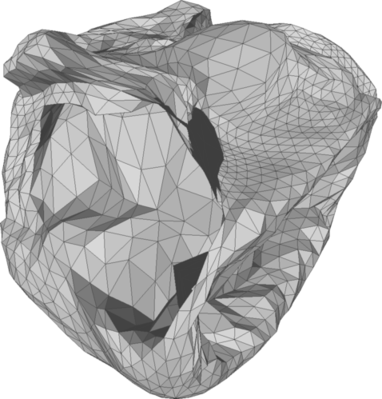} &
		\includegraphics[width=0.08\textwidth]{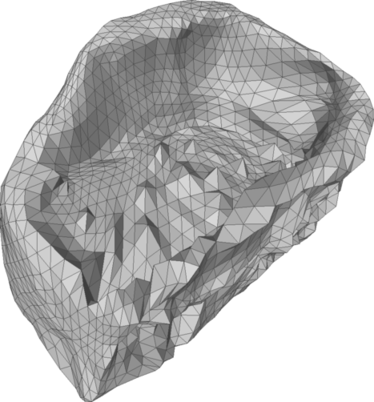}  &
		\includegraphics[width=0.03\textwidth]{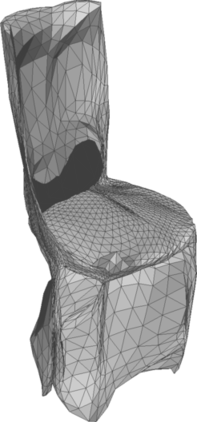}&
		\includegraphics[width=0.08\textwidth]{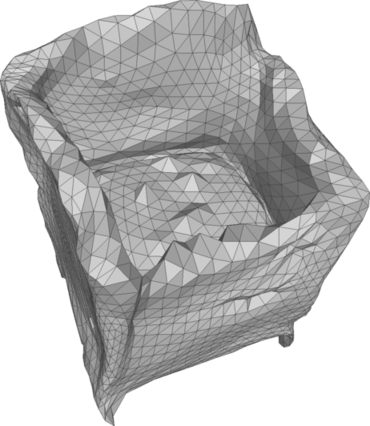} 
		\\
		\includegraphics[width=0.065\textwidth]{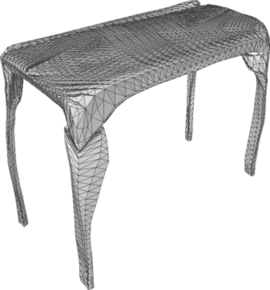} &
		\includegraphics[width=0.08\textwidth]{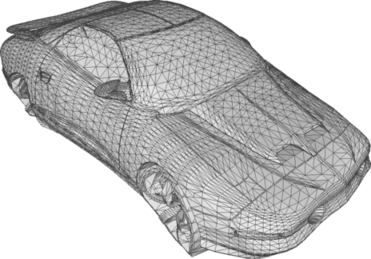}  &
		\includegraphics[width=0.08\textwidth]{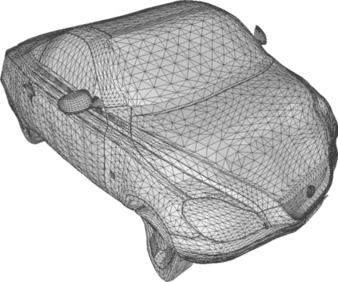} &
		\includegraphics[width=0.08\textwidth]{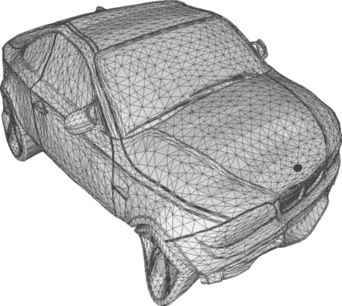} &
		\includegraphics[width=0.03\textwidth]{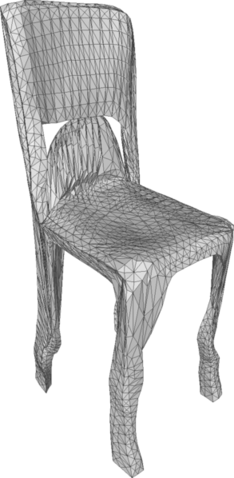} &
		\includegraphics[width=0.065\textwidth]{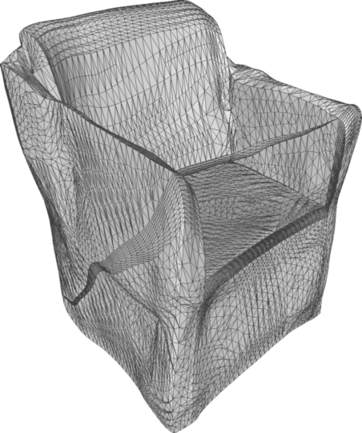}
	\end{tabular}
	\caption{Qualitative results on online product images. The first row shows the images scrapped online. Second and third row are results of AtlasNet and Pixel2Mesh respectively. Last row is our results.}
	\vspace{-5pt}
	%\vspace{-3mm}
	\label{fig:2dreal}
\end{figure}

\vspace{-10pt}
\paragraph{Evaluation on real images.} We further evaluate our method on real product images that can be found online. For each input image, we select a source model as described before and provide the deformation result. Even though our method has been trained only on synthetic images, we observe that it generalizes to real images as seen in Figure~\ref{fig:2dreal}. AtlasNet and Pixel2Mesh fail in most cases, while our method is able to generate plausible results by taking advantages of source meshes.

\subsection{Ablation Study}
\label{sec:ablation}
We study the importance of different losses and the offset decoder architecture on ShapeNet chair category. We compare our final model to variants including 1) 3DN without the symmetry loss, 2) 3DN without the mesh Laplacian loss, 3) 3DN without the local permutation invariance loss, and 4) fusing global features with midlayer features instead of the original point positions (see the supplemental material for details).

We provide quantitative results in Table~\ref{table:losses}. Symmetry loss helps the deformation to produce plausible symmetric shapes. Local permutation and Laplacian losses help to obtain smoothness in the deformation field across 3D space and along the mesh surface. However, midlayer fusion makes the network hard to converge to a valid deformation space.

\begin{table}
	\begin{center}
		\setlength{\tabcolsep}{0.1em}
		\newcolumntype{C}{>{\centering\arraybackslash}p{2.5em}}
		\begin{tabular}{c|ccccc}
			\Xhline{3\arrayrulewidth}
			&CD&EMD&IoU\\
			\hline
			3DN& \textbf{4.50} &\textbf{2.06} & \textbf{41.0} \\
			-Symmetry&4.78&2.73 & 36.7\\
			-Mesh Laplacian& 4.55&2.08& 39.8\\
			-Local Permutation& 5.31&2.96& 35.4\\
			Midlayer Fusion& 6.63&3.03& 30.9\\
			\Xhline{3\arrayrulewidth}
		\end{tabular}
	\end{center}
	\hspace{-6pt}
	\caption{Quantitative comparison on ShapeNet rendered images. '-x' denotes without x loss. Metrics are CD ($\times 1000$), EMD ($\times 0.01$) and IoU (\%).}
	\label{table:losses}
	%\vspace{-20pt}
	%\vspace{-3mm}
\end{table}

%\begin{figure}
%	\centering
%	\renewcommand{\arraystretch}{0.1}% Tighter
%	\begin{tabular}{cccc}
%
%		\includegraphics[height=0.1\textwidth]{figures/ablation/local} &
%		\includegraphics[height=0.1\textwidth]{figures/ablation/midfusion} &
%		\includegraphics[height=0.1\textwidth]{figures/ablation/symmetry} &
%		\includegraphics[height=0.1\textwidth]{figures/ablation/laplacian} 
%		\\
%		(a)&(b)&(c)&(d)
%		
%	\end{tabular}
%	\caption{Qualitative results on ShapeNet rendered images.}
%	%\vspace{-3mm}
%	\label{fig:losses}
%\end{figure}

\subsection{Applications}
\begin{figure}
 	\centering
 	\renewcommand{\arraystretch}{0.1}% Tighter
 	\newcolumntype{C}{>{\centering\arraybackslash}p{3em}}
 	\begin{tabular}{C|CCCC}
 		\diagbox[width=4em]{S}{T}&
 		\includegraphics[height=0.04\textwidth]{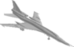} &
 		\includegraphics[height=0.04\textwidth]{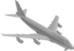} &
 		\includegraphics[height=0.04\textwidth]{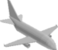} &
 		\includegraphics[height=0.04\textwidth]{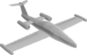} 
 		\\
 		\hline
 		\\
 		\includegraphics[height=0.04\textwidth]{figures/gridplane/0_0_src.png} &
 		\includegraphics[height=0.04\textwidth]{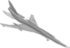} &
 		\includegraphics[height=0.04\textwidth]{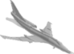} &
 		\includegraphics[height=0.04\textwidth]{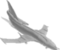} &
 		\includegraphics[height=0.04\textwidth]{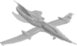} 
 		\\
 		\includegraphics[height=0.04\textwidth]{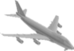} &
 		\includegraphics[height=0.04\textwidth]{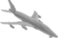} &
 		\includegraphics[height=0.04\textwidth]{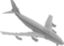} &
 		\includegraphics[height=0.04\textwidth]{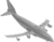} &
 		\includegraphics[height=0.04\textwidth]{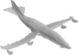} 
 		\\
 		\includegraphics[height=0.04\textwidth]{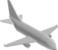} &
 		\includegraphics[height=0.04\textwidth]{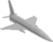} &
 		\includegraphics[height=0.04\textwidth]{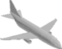} &
 		\includegraphics[height=0.04\textwidth]{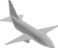} &
 		\includegraphics[height=0.04\textwidth]{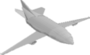} 
 		\\
 		\includegraphics[height=0.04\textwidth]{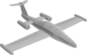} &
 		\includegraphics[height=0.04\textwidth]{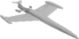} &
 		\includegraphics[height=0.04\textwidth]{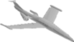} &
 		\includegraphics[height=0.04\textwidth]{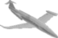} &
 		\includegraphics[height=0.04\textwidth]{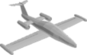} 
 		
 	\end{tabular}
 	\caption{Deformation with different source-target pairs. `S' and `T' denote source meshes and target meshes respectively.}
 	%\vspace{-3mm}
 	\label{fig:shapegensame}
 \end{figure}

\paragraph{Random Pair Deformation.} In Figure~\ref{fig:shapegensame} we show deformation results for randomly selected source and target model pairs. While the first column of each row is the source mesh, the first row of each column is the target. Each grid cell shows deformation results for the corresponding source-target pair.
 
\begin{figure}
	\centering
	\renewcommand{\arraystretch}{0.1}% Tighter
	\newcolumntype{C}{>{\centering\arraybackslash}p{1.7em}}
	\begin{tabular}{c|CCCCCCC}
		\includegraphics[height=0.04\textwidth]{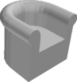} &
		\includegraphics[height=0.04\textwidth]{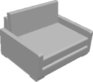} &
		\includegraphics[height=0.04\textwidth]{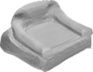} &
		\includegraphics[height=0.04\textwidth]{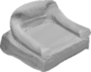} &
		\includegraphics[height=0.04\textwidth]{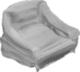} &
		\includegraphics[height=0.04\textwidth]{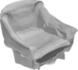} &
		\includegraphics[height=0.04\textwidth]{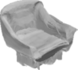} &
		\includegraphics[height=0.04\textwidth]{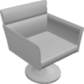} 
		\\
		\includegraphics[height=0.04\textwidth]{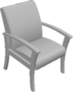} &
		\includegraphics[height=0.04\textwidth]{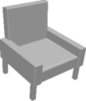} &
		\includegraphics[height=0.04\textwidth]{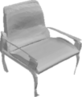} &
		\includegraphics[height=0.04\textwidth]{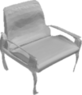} &
		\includegraphics[height=0.04\textwidth]{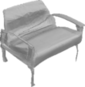} &
		\includegraphics[height=0.04\textwidth]{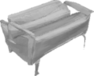} &
		\includegraphics[height=0.04\textwidth]{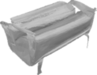} &
		\includegraphics[height=0.04\textwidth]{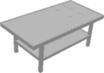} 
		\\
		\includegraphics[height=0.04\textwidth]{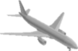} &
		\includegraphics[height=0.04\textwidth]{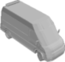} &
		\includegraphics[height=0.04\textwidth]{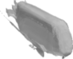} &
		\includegraphics[height=0.04\textwidth]{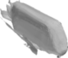} &
		\includegraphics[height=0.04\textwidth]{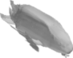} &
		\includegraphics[height=0.04\textwidth]{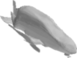} &
		\includegraphics[height=0.04\textwidth]{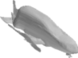} &
		\includegraphics[height=0.04\textwidth]{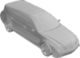} 
		\\
		Source& Target1 &\multicolumn{5}{c}{Interpolation}&Target2
	\end{tabular}
	\caption{Shape interpolation.}
	%\vspace{-10pt}
	%\vspace{-3mm}
	\label{fig:shapeinterpolation}
\end{figure}

 \vspace{-10pt}
\paragraph{Shape Interpolation.} Figure~\ref{fig:shapeinterpolation} shows shape interpolation results. Each row shows interpolated shapes generated from the two targets and the source mesh. Each intermediate shape is generated using a weighted sum of the global feature representations of the target shapes. Notice how the interpolated shapes gradually deform from the first to the second target.
\vspace{-10pt}

\begin{figure}
	\centering
	\renewcommand{\arraystretch}{0.02}% Tighter
	\newcolumntype{C}{>{\centering\arraybackslash}p{1.9em}}
	\hspace{-10pt}
	\begin{tabular}{CCCCCCCC}
		
		\includegraphics[width=0.03\textwidth]{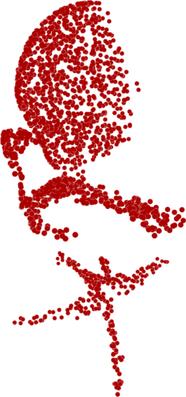} &
		\includegraphics[width=0.04\textwidth]{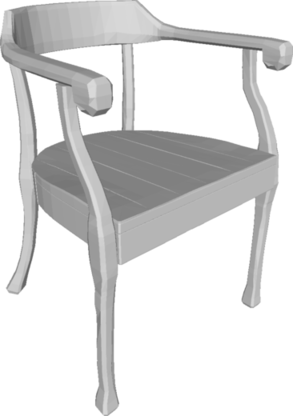} &
		\includegraphics[width=0.03\textwidth]{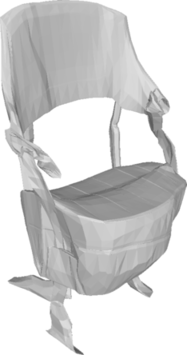} &
		\includegraphics[width=0.04\textwidth]{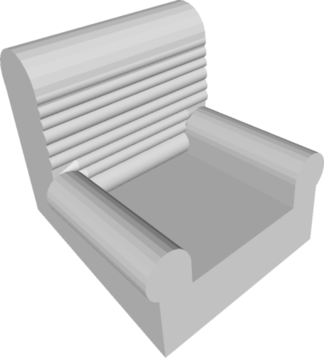} &
		\includegraphics[width=0.04\textwidth]{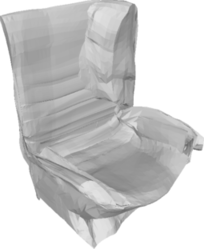} &
		\includegraphics[width=0.04\textwidth]{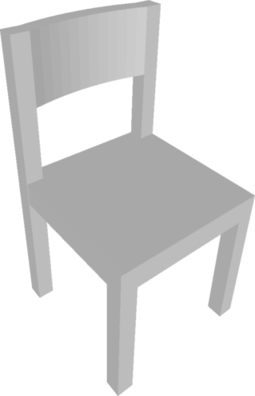} &
		\includegraphics[width=0.04\textwidth]{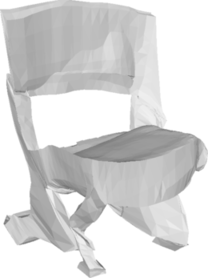} &
		\includegraphics[width=0.04\textwidth]{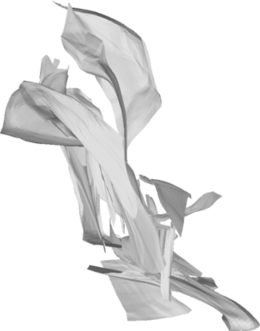} 
		\\
		\includegraphics[width=0.04\textwidth]{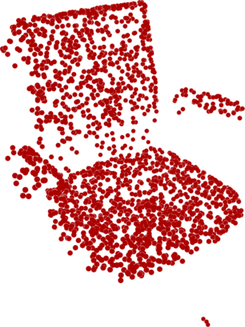} &
		\includegraphics[width=0.04\textwidth]{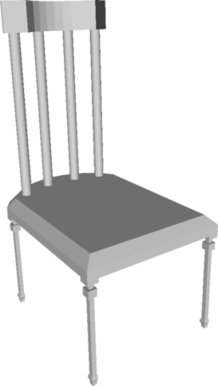} &
		\includegraphics[width=0.04\textwidth]{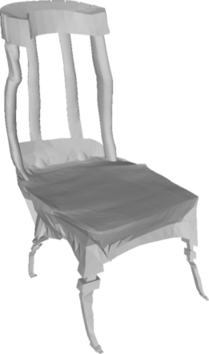} &
		\includegraphics[width=0.04\textwidth]{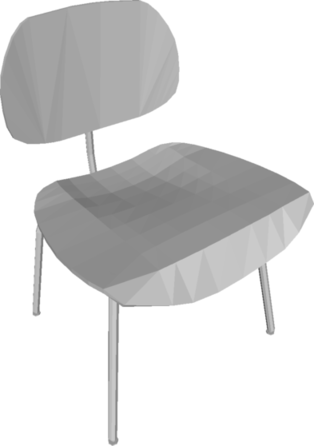} &
		\includegraphics[width=0.04\textwidth]{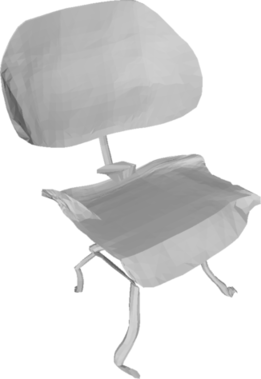} &
		\includegraphics[width=0.04\textwidth]{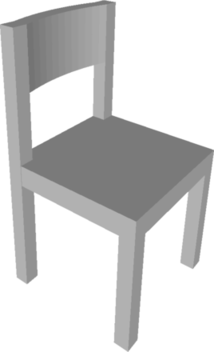} &
		\includegraphics[width=0.04\textwidth]{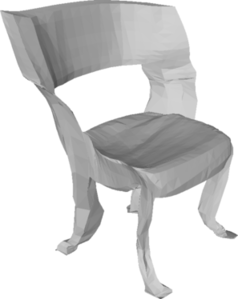} &
		\includegraphics[width=0.04\textwidth]{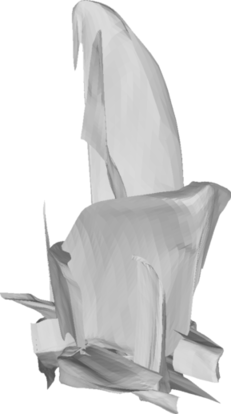} 
		\\
		\includegraphics[width=0.04\textwidth]{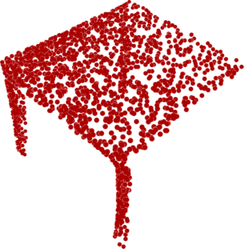} &
		\includegraphics[width=0.04\textwidth]{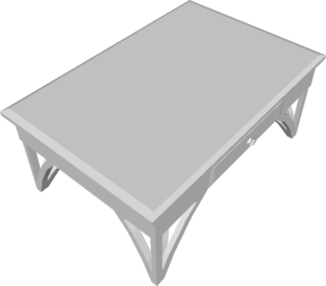} &
		\includegraphics[width=0.04\textwidth]{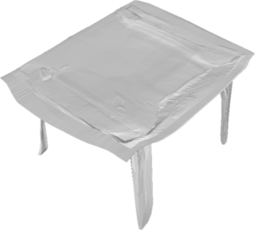} &
		\includegraphics[width=0.04\textwidth]{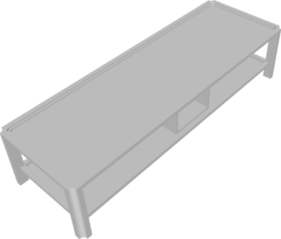} &
		\includegraphics[width=0.04\textwidth]{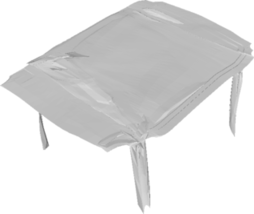} &
		\includegraphics[width=0.04\textwidth]{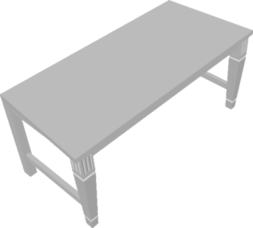} &
		\includegraphics[width=0.04\textwidth]{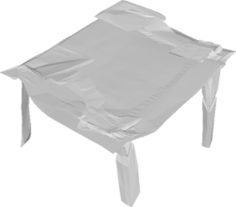} &
		\includegraphics[width=0.04\textwidth]{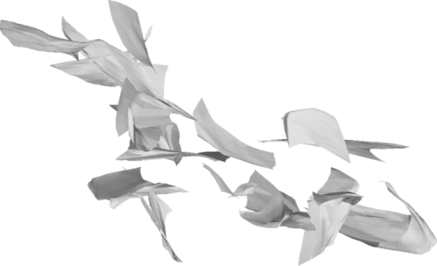} 
		\\
		\small{Scan}&\small{Src1}&\small{Out1}&\small{Src2}&\small{Out2}&\small{Src3}&\small{Out3}&\small{AtlasNet}
	\end{tabular}
	\caption{Shape inpainting with real point cloud scan as input. Src means source mesh and 'out' is the corresponding deformed mesh.}
	\vspace{-10pt}
	%\vspace{-3mm}
	\label{fig:realpointcloud}
\end{figure}

\paragraph{Shape Inpainting.} We test our model trained in Section~\ref{sec:3d} on targets in the form of partial scans produced by RGBD data~\cite{Sung:2015}. We provide results in Figure~\ref{fig:realpointcloud} with different selection of source models. We note that AtlastNet fails on such partial scan input.
\vspace{-5pt}

%\paragraph{Part Correspondences}

%% file: conclusion.tex
\section{Conclusion}
We have presented 3DN, an end-to-end network architecture for mesh deformation. Given a source mesh and a target which can be in the form of a 2D image, 3D mesh, or 3D point clouds, 3DN deforms the source by inferring per-vertex displacements while keeping the source mesh connectivity fixed. We compare our method with recent learning based surface generation and deformation networks and show superior results. Our method is not without limitations, however. Certain deformations indeed require to change the source mesh topology, e.g., when deforming a chair without handles to a chair with handles. If large holes exist either in the source or target models, Chamfer and Earth Mover's distances are challenging to compute since it is possible to generate many wrong point correspondences.

In addition to addressing the above limitations, our future work include extending our method to predict mesh texture by taking advantages of differentiable renderer~\cite{kato2018renderer}.